\documentclass[runningheads]{llncs}

\usepackage{amsmath,amssymb}
\usepackage[dvipsnames]{xcolor}

\usepackage{graphbox}
\usepackage{multirow}
\usepackage{makecell}
\usepackage{graphicx}
\usepackage{booktabs}
\usepackage{pifont}
\newcommand{\cmark}{\ding{51}}%
\newcommand{\xmark}{\ding{55}}%

\makeatletter
\renewcommand{\paragraph}{%
  \@startsection{paragraph}{4}%
  {\z@}{0em}{-1em}%
  {\normalfont\normalsize\bfseries}%
}
\makeatother

\DeclareMathOperator*{\argmax}{arg\,max}

\definecolor{red}{rgb}{0.95,0.4,0.4}
\definecolor{green}{rgb}{0.55,1.0,0.55}
\definecolor{lightgreen}{rgb}{0.75,1.0,0.75}
\definecolor{blue}{rgb}{0.4,0.4,0.95}
\definecolor{darkblue}{rgb}{0,0,0.8}
\definecolor{darkred}{rgb}{0.8,0,0}
\definecolor{darkgreen}{rgb}{0,0.5,0}
\definecolor{grey}{rgb}{0.6,0.6,0.6}
\definecolor{amber}{RGB}{255,210,43}

\usepackage[export]{adjustbox}

\usepackage{tikz}

\usetikzlibrary{math}
\usetikzlibrary{fit}

\tikzset{%
nodeA/.style={circle,draw,black,fill=blue,inner sep=0pt,minimum size=6pt},
nodeB/.style={rectangle,draw,black,fill=orange,inner sep=0pt,minimum size=6pt},
nodeC/.style={circle,draw,white,fill=white,inner sep=0pt,minimum size=0pt},
container/.style={draw, dashed, rectangle, inner sep=0.3cm, rounded corners}
}

\definecolor{myred}{rgb}{0.7,0.25,0.2}
\definecolor{mygreen}{RGB}{0, 153, 0}
\definecolor{myblue}{RGB}{0, 153, 255}
\definecolor{myorange}{RGB}{255, 153, 51}
\definecolor{mygray}{rgb}{0.4,0.4,0.4}

\usepackage{graphicx}
\usepackage{comment}

\usepackage{color}
\usepackage{url}
\usepackage{hyperref}
\usepackage{cleveref}

\usepackage{hyperref}
\usepackage{academicons}
\usepackage{xcolor}
\usepackage{orcidlink}

\newcommand{\orcid}[1]{\href{https://orcid.org/#1}{\textcolor[HTML]{A6CE39}{\aiOrcid}}}

\newif\ifreview
\reviewfalse

\ifreview
	\usepackage{lineno}

	\linenumbers
\fi

\begin{document}


\setlength{\abovedisplayskip}{3pt}
\setlength{\belowdisplayskip}{5pt}
\setlength{\abovedisplayshortskip}{3pt}
\setlength{\belowdisplayshortskip}{5pt}

\def\SubNumber{1}

\def\GCPRTrack{Special Track: Pattern recognition in the life and natural sciences}

\title{Animal Identification with Independent Foreground and Background Modeling}




\ifreview
\else
        \titlerunning{Animal Id. with Independent Foreground and Background Modeling}

	\author{Lukas Picek\inst{1,2}\orcidlink{0000-0002-6041-9722} \and
	Lukáš Neumann\inst{3}\orcidlink{0000-0002-9428-3712} \and
	Jiří Matas\inst{3}\orcidlink{0000-0003-0863-4844}}
	
	\authorrunning{L. Picek et al.}
	
	\institute{
 University of West Bohemia, FAS, Department of Cybernetics, Pilsen, Czechia \and 
 Inria, LIRMM, University of Montpellier, CNRS, Montpellier, France\\ \and 
 Czech Technical University in Prague, FEL, CMP, Prague, Czechia\\
	\email{lukaspicek@gmail.com, lukas.neumann@cvut.cz, matas@fel.cvut.cz}}
\fi

\maketitle              

\begin{abstract}
We propose a method that robustly exploits background and foreground in visual identification of individual animals.
Experiments show that their automatic separation,
made easy with methods like Segment Anything, together with independent foreground and background-related modeling, improves results.
The two predictions are combined in a principled way, thanks to 
novel Per-Instance Temperature Scaling that helps the classifier
to deal with appearance ambiguities in training and to produce calibrated outputs in the inference phase. 
For identity prediction from the background, we propose novel spatial and temporal models. 
On two problems, the relative error w.r.t. the baseline was reduced by 22.3\% and 8.8\%, respectively.
For cases where objects appear in new locations, an example of background drift, accuracy doubles. 
\end{abstract}    
\section{Introduction}\vspace{-0.15cm}
\label{sec:intro}
Both the foreground and the background play a role in visual object recognition and identification \cite{lauer2021role,oliva2007role}; their relative contributions to the final decision vary. It is easy to imagine situations from both ends of the spectrum. 
At the one end, the foreground captures all relevant information in a passport photograph with a standard uniform background. At the other, one might not identify the neighbor's dachshund if it were not for the face of the owner walking it.
The same holds for categorization -- very few people would assume the small yellowish object 
on a tennis
court to be an apple, unlike the similarly 
looking object 
in a supermarket, 
at least when viewed from a distance or in poor lighting.

\begin{figure}
\centering
\begin{tabular}{c@{\hspace{3px}}cc}

    \includegraphics[height=2.3cm,width=3.5cm,align=c]{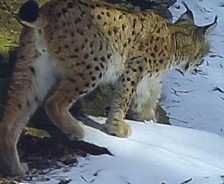} \vspace{2pt} & 
    \includegraphics[height=2.3cm,width=3.5cm,align=c]{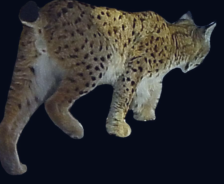} \vspace{2pt} &  \includegraphics[height=2.3cm,width=3.5cm,align=c]{figures/gradcam/fgsam/MsB_15120_rgb.png} \hspace{-58pt}\raisebox{-25.0pt}{\colorbox{white}{\scriptsize + prior $\Psi_k(\theta)$}} \\
     \includegraphics[height=2.3cm,width=3.5cm,align=c]{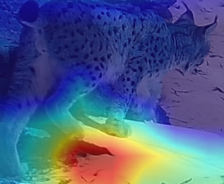} \vspace{2pt} & {\hspace{-1mm}}\includegraphics[height=2.3cm,width=3.5cm,align=c]{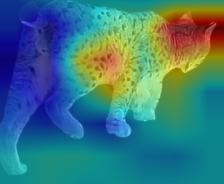}\vspace{2pt} & {\hspace{1px}}\includegraphics[height=2.3cm,width=3.5cm,align=c]{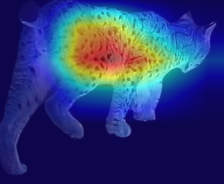} \\
     49.71\% & 49.26\% & \textbf{60.89\%}

\end{tabular}
\vspace{-0.1cm}
\caption{
\textbf{Left:} animal re-identification often relies on background pixels correlated with an identity frequenting a small set of locations. \textbf{Center:} removing the background ensures focus on the individual’s appearance but might reduce accuracy. \textbf{Right:} adding background in the form of prior $\Psi_k(\theta)$ and \textit{calibrating the network} for each observation individually alleviates the problem and significantly improves accuracy.
}
\vspace{-0.35cm}
\label{fig:splash}
\end{figure}
This paper proposes\footnote{Source code is available at \url{https://github.com/BohemianVRA/frg-bgr-modeling}.} a new method for visual identification that robustly exploits both background and foreground.
Most computer vision methods are dominantly based on machine learning, trained on data assumed to be from the same distribution as the data encountered at deployment (test) time \cite{liu2021swin,liu2022convnet,oquab2023dinov2}.
Yet real test data are prone to distribution and domain shifts, which are often different for the foreground and the background, where it is more prominent, e.g., when seeing an object in a never-seen-before context, in a new setting \cite{kondrateva2021domain,picek2022automatic,sun2022shift}. 
As an example of a particular instance of such shift, consider the model of a car that is strongly coupled with the dominant background, the road. 
Photos of a car in a swimming pool or on the moon would probably not be recognized.
On the other hand, the background may provide probabilistic information about the foreground class via, e.g., its semantic class (day/night, underwater, etc.),
a particular location, or time. In wildlife conservation, where individuals are difficult to recognize, the background provides strong disambiguation cues for individuals from many species, see \Cref{fig:splash}.
Handling the foreground and background separately, made possible by the publishing of the Segment Anything Model (SAM) \cite{kirillov2023segany}, thus has significant potential.

In this paper, we demonstrate the interplay of the foreground and background in the wildlife re-identification problem because the similar appearance of individuals hinders recognition to the point that, for some species, it is very challenging to recognize individuals just from the foreground, and proper handling of the information related to the background has a significant impact. The long-term nature of the monitoring implies inherent changes to individuals' appearance, location, and surroundings due to seasonal and vegetation changes, which have very different impacts on the foreground and the background. \vspace{-0.2cm} \\

\noindent
\textbf{The main contributions of the paper include}:
\begin{enumerate}
\vspace{-0.25cm}
    \item A novel robust method for background information modeling. The background and foreground predictions are combined in a principled way. 
    \item  A novel Per-Instance Temperature Scaling method -- PITS -- that improves the accuracy of the fine-grained recognition on its own; when combined with explicit prior modeling, we show the accuracy is significantly improved.     
    Combining foreground and background predictions requires classifier calibration.   
    \item We show that explicit prior modeling improves recognition accuracy, especially for images where the object appears at a new location.
\end{enumerate}
\section{Related Work}
\label{sec:relatedwork}
\vspace{-0.2cm}
\paragraph{Animal re-identification.}
Image-based animal re-identification is about assigning an “identity” to an animal based on unique visual characteristics (e.g., spots or earth shape). The process is essential for various aspects of ecology, including but not limited to studying wildlife populations, movements, behavior, and management \cite{hayden2022automated,joska2021acinoset,Ng_2022_CVPR,papafitsoros_2021,Schofield_2022,tuia2022perspectives,vidal2021perspectives}. 
However, the continuously growing image datasets from long-lasting projects highlight the need for automated methods \cite{Schofield_2020,swanson2015snapshot}. The commonly used approaches include 
\textit{(i) Species-specific methods:} Tailored to individual species or related groups, focusing on unique visual features, often with manual preprocessing \cite{bedetti2020system,drechsler2015genetic,gilman2016computer,weideman2020extracting},
\textit{(ii) Local descriptors:} Identify local keypoints and extract descriptors without fine-tuning; popular systems include HotSpotter and WildID \cite{andrew2016automatic,dunbar2021hotspotter,reno2019sift}, and
\textit{(iii) Deep descriptors:} Use neural networks to create image representations for matching; performance improves with models like BioCLIP, DINOv2, and MegaDescriptor \cite{bruslund2020re,vcermak2024wildlifedatasets,deb2018face,miele2021revisiting,ueno2022automatic}.

\vspace{0.2cm}
\paragraph{Foreground--background separation.}
Classifiers often focus on background over foreground features, especially with limited data, and methods usually address this by emphasizing foreground features, potentially suppressing useful background information. This sensitivity was extensively studied \cite{barbu2019objectnet,bhatt2024mitigating,moayeri2022comprehensive,shetty2019not,xiao2020noise}.
Bhatt et al. \cite{bhatt2024mitigating} explored part-based learning to separate foreground and background, reducing classifier sensitivity to weakly annotated background parts by favoring foreground features with a dedicated loss function. This improves foreground representation but may restrict learning strong background features.
Barbu et al. \cite{barbu2019objectnet} and Xiao et al.\cite{xiao2020noise} address training dataset bias by mitigating correlations between the foreground and the background. ObjectNet \cite{barbu2019objectnet} provides a bias-controlled dataset, while Xiao et al. \cite{xiao2020noise} uses random backgrounds for foreground object placement. 
Similarly, Shetty et al. \cite{shetty2019not} propose object-removal augmentation. 
These methods assume that the background distribution should be uninformative; while random backgrounds may render them irrelevant, in many challenging foreground classification scenarios, the background remains beneficial, as shown in this study. 
\textit{None of the methods explicitly model the relationship between a background and the animal's identity.} 

\vspace{0.2cm}
\paragraph{Classifier calibration.}
The most common methods adjust a trained classifier's prediction scores using parameters derived from a validation set that matches the training data's distribution but wasn't used in training.
Guo et al. \cite{guo2017calibration} introduced temperature scaling to the softmax function, adding a global temperature parameter to reduce overconfidence. Wenger et al. \cite{wenger2020non} used a latent Gaussian process for multi-class classification, but its complexity is impractical for 400+ identities. Neumann et al.~\cite{neumann2018relaxed} allowed different temperature values for each training sample but still required post-processing scaling.
Regularization methods modify the training objective for calibration. Szegedy et al. \cite{szegedy2016rethinking} introduced label smoothing, which softens the hard class labels into a distribution over labels, reducing the model's confidence in any single label. Pereyra et al.~\cite{pereyra2017regularizing} added negative entropy terms to penalize overconfidence. Thulasidasan et al. \cite{thulasidasan2019mixup} used mixup data augmentation, interpreted as entropy-based regularization.
%

\section{Method}
\label{sec:method}
We pose animal identification (i.e., animal re-identification) as a classification problem where each individual animal identity is represented by its own class. In a typical setting~\cite{seaturtleid2022,vcermak2024wildlifedatasets}, the training set is defined by a time constraint, i.e., all individuals observed up to a given date (typically the end of a season), and a testing set is based on new observations from an upcoming season.

In such classification setup, Cross-Entropy (CE) loss is typically used to train a deep neural network $\Phi(x)$, where
its output with an appropriate normalization of its outputs, such as temperature-scaled softmax activation $\sigma(z)$, can be treated as an estimate of  the posterior probability $p_T(k | x)$ of class $k$ (in our case identity) given observation~$x$

\begin{equation}
    p_T(k | x) = \frac{p_T(x | k)\,p_T(k)}{p_T(x)} \approx \sigma\big(\Phi(x)\big),
    \label{eq:posterior}
    \vspace{0.1cm}
\end{equation}

since $p_T(k | x)$ is the minimizer of the CE loss and temperature scaling aims at making the approximation close for the top-1 class, calibrating the classifier~\cite{guo2017calibration}. The subscript $T$ highlights the fact that the approximation is valid for the distribution if the training set $T$ was drawn from. It is quite common that observations $x$ are drawn from different distributions at deployment $D$ (test) time due to various domain shifts. The prior $p_T(k)$ is often uniform while at deployment, classes have a long-tail distribution $p_D(k)$.

In experiments validating our approach, the visual recognition task is the identification of an individual. Since our training and evaluation data are split by time, $p_T(x|k) \neq p_D(x|k)$ as every individual changes its appearance with age. Nevertheless, we ignore the difference and treat $p_T(x|k) = p_D(x|k) = p(x|k)$, avoiding the need to deal with test time adaptation, which is not possible in the limited space. Ignoring this concept drift reduces performance but does not interfere with our proposed foreground-background separation handling.  

When training on the whole image, the class (identity) prior $p_T(k)$ in the training set is inherently incorporated into the network $\Phi(x)$ during training.
As shown in \Cref{fig:splash}, a standard classification network $\Phi(x)$ actually exploits the image background to estimate $p(k | x)$.  We posit that the image background, in different ways for different visual recognition tasks,  defines the prior probability $p(k)$ and should not be used to model an object's (here individual) appearance. Thus, 
we force the network to focus only on the appearance of the identity, i.e., by using only image pixels that belong to the animal to be identified (denoted as foreground $x_{\text{FG}}$).
\begin{equation}
    p(x | k) \approx \Phi(x_{\text{FG}})
    \vspace{0.1cm}
\end{equation}
We then introduce a simple model $\Psi_k$ for the class (identity) prior $p(k)$.
This model can be estimated using data from the background or supplemented with additional information, such as the location and date of image capture, or a combination of both.


\noindent \textbf{Therefore, we define the prior model $\Psi_k$ as}
\begin{equation}  
    p(k) \approx \Psi_k(x_{\text{BG}}, x_\theta), \label{eq:ourprior}
\end{equation}

where $x_{\text{BG}}$ is the image background and $x_\theta$ represents additional measurements available in the image metadata that might affect the prior, such as the time or the location of the observation.
Our final model for classification (re-identification) is then expressed as
\begin{equation}
    p(k\,|\,x_{\text{FG}}, x_{\text{BG}}, x_\theta) = \frac{\Phi(x_{\text{FG}})\,\Psi_k(x_{\text{BG}}, x_\theta)}{Z}, \label{eq:ourposterior}
\end{equation}

where $Z$ is a normalization factor (the factor $Z$ can be omitted, as its value does not affect which class comes out with the highest score).

\subsection{Foreground Model} 
\label{sec:pitsloss}
Modern deep neural networks are systematically over-confident in their predictions~\cite{guo2017calibration}. If their predictions are interpreted as the posterior probability $p(k|x)$ (see \Cref{eq:posterior}), and one is only interested in the class $k$ with the highest probability -- which is the case in most common classification tasks -- this does not pose a problem. In our case, we want the network predictions to be interpreted as the likelihood $p(x|k)$ and to infer the posterior explicitly by multiplying the likelihood with the identity prior $p(k)$ (see \Cref{eq:ourposterior}), for which it is paramount that the network output $\Phi(x)$ is calibrated, in other words the network output should be as close to the probability distribution $p(x|k)$ as possible. \vspace{-0.1cm} \\

\paragraph{Temperature Scaling.} The most common approach to deep network calibration is temperature scaling~\cite{guo2017calibration}, which introduces an additional global \textit{temperature} parameter $T$ into the final softmax layer $\sigma(\mathbf{z_i})$

\begin{equation} 
 {\sigma}_i(\mathbf{z_i}, T) = \frac{\exp\left(\frac{\mathbf{z}_i}{T}\right)}{\sum^K_{k=1} \exp\left(\frac{\mathbf{z}^k_i}{T}\right)},
  \label{eq:temperaturescaling}
\end{equation}

where $\mathbf{z}_i$ are the outputs (logits) from an already trained deep network $\Phi(x)$.
The parameter value $T$ is found as a post-processing step for an already trained network $\Phi(x)$ by minimizing the standard CE loss, while \textit{keeping the weights of the original network fixed}, i.e., $\mathcal{L}(\mathbf{z_i},T,y_i) = -\sum_{i=1}^N \log \sigma^{y_i}_i(z_i, T)$. \vspace{-0.1cm} \\

\begin{figure}[t]
    \centering
    
    \begin{tabular}{c@{\hspace{7.5pt}}|@{\hspace{7.5pt}}c} 
    
        \adjustbox{valign=t}{\includegraphics[width=5.7cm]{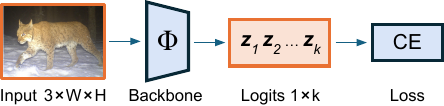}} &         
        \adjustbox{valign=t}{\includegraphics[width=5.7cm]{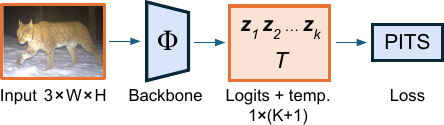}} \\  
    \end{tabular}
    
    \caption{
\textbf{Left:} Standard approach with Cross-Entropy loss. \textbf{Right:} PITS scaling to $T_i$.}   
    \label{fig:pipeline}
    \vspace{-10pt}
\end{figure}

\paragraph{Per-Instance Temperature Scaling (PITS).} 
In our method, we modify the temperature scaling paradigm above and use the deep neural network $\Phi(x)$ to output the temperature scaling factor $T_i$ for every sample $i$ directly -- the network outputs the logits vector $\mathbf{z}_i$ and one additional scalar $T_i$ (see \Cref{fig:pipeline}). The output of the network is then given again by a modified softmax layer

\begin{equation} 
 {\sigma}_i(\mathbf{z_i}, T_i) = \frac{\exp\left(\frac{\mathbf{z}_i}{T_i}\right)}{\sum^K_{k=1} \exp\left(\frac{\mathbf{z}^k_i}{T_i}\right)}, \quad T_i \geq 1
  \label{eq:perinstancetemperaturescaling}
\end{equation}

This change allows the network to adjust temperatures based on sample sizes. For identities with fewer training samples, a higher temperature 
$T_i$
is preferred to reflect uncertainty in decisions. Conversely, for identities with hundreds of samples, 
$T_i$
can be closer to 1, indicating greater confidence due to sufficient training data.
More formally, the network $\Phi$ outputs a vector of logits $\mathbf{z}_i$ and scalar $T_i$ for a sample $x_i$. The network is trained by optimizing the loss
\begin{align}
\mathcal{L}(x_i,y_i) &= \sum_{i=1}^N \Big(-\log \sigma^{y_i}_i(\mathbf{z}_i, T_i) + \lambda\,\lVert T_i - \mathcal{T}({y_i}) \rVert^2 \Big).\label{eq:finalloss} 
\end{align}
The network is therefore optimized to output correct classification as in the standard CE loss, but the network can soften this requirement for ambiguous samples by setting high temperature $T_i$, but at the same time, the temperature $T_i$ is incentivized to stay close to the desired class-specific temperature $\mathcal{T}(k)$.

Our method uses the following formula to model that classes (identities) with a low number of training samples should have a high per-instance temperature 
\begin{equation} 
\mathcal{T}(k) = 1 - \log \frac{N_k}{\max_{k` \in K}N_{k'}},
\end{equation}

where $N_k$ denotes the number of class (identity) training samples $k$. Note that for balanced datasets where each class (identity) has a similar number of training samples ($N \approx N_k\;\forall k$), $\mathcal{T}(k)$ is 1 and the \Cref{eq:perinstancetemperaturescaling} is reduced to the CE loss. For long-tailed class distributions, $\mathcal{T}(k)$ still remains close to 1 for frequent classes, while for the rare classes, the value $\mathcal{T}(k)$ is high, allowing the network not to make hard decisions for rare instances.  \vspace{-0.05cm}

\subsection{Background Model}
\label{sec:backgroundmodel}

In \Cref{eq:ourprior}, our background model determines the prior probability for each class (identity) 
$k$, leveraging information from the image background 
$x_{\text{BG}}$
or image metadata 
$x_\theta$.
This model is tailored to the task and varies with the specific domain; our paper introduces three background models (priors) relevant to animal re-identification. Different domains may necessitate alternative background models, which we defer to future research beyond the scope of this paper.  \vspace{-0.1cm}\\

\paragraph{Home Location (HL) Prior.} We hypothesize that the geographical location where the object was captured plays a substantial role in estimating the class (identity) prior to $p(k)$. The geographical location is implicitly encoded in the image background $x_{\text{BG}}$, or it can be available in image metadata $x_\theta$ as a GPS coordinate.
For image $x_i$, we denote the camera geographical location as $l_i$ and formulate the \textit{Home Location Prior} as

\begin{equation}
    \Psi_k(H_k, l_i) = \frac{
         \exp\big(-\alpha\,\lVert H_k - l_i \rVert_{2} \big) 
       }{Z},
    \label{eq:homelocation}
\end{equation}

where $H_k$ denotes the most frequent location of the identity $k$ in the training set (i.e., the ``home'' location), $\alpha$ is a decay parameter whose is determined empirically on the training/validation set, and $Z$ is a normalization factor in ensuring the distribution sums to 1. 

We present two versions of the prior: \textbf{(i) $\text{HL}_{\text{BG}}$} prior exploits image background to determine camera location by using a deep network $\Phi'(x)$ trained on the training set to classify the background image and to select the camera location from a finite set of possible locations. \textbf{(ii) $\text{HL}_\theta$ prior}, on the other hand, uses image GPS metadata to select the image location. The set of possible locations is, in both cases, represented by the same geospatial grid, which covers the whole region of interest, with reasonably sized grid cells, for example, $5\times5$\,\,km.  \vspace{-0.1cm}\\

\paragraph{Migrating Location (ML) Prior. } The assumption of a single ``home'' location is not fully realistic because, indeed, in the domain of identity re-identification, a given identity can move around, changing its geographical location. At the same time, physical constraints apply, and it is, of course, not possible for one individual to be captured by two different cameras 80km apart within the span of 10 minutes. 
To allow for gradual movement while still penalizing for impossible ``jumps'' between locations, we soften the Home Location (HL) prior of \Cref{eq:homelocation} and introduce the \textit{Migrating Location  Prior} as
\begin{eqnarray}
    \Psi_k(\mathbf{L}_k, l_i) &= & \frac{\exp\big(-\alpha\,\lVert  \mathbf{L}_k - l_i \rVert_2 \big)}{Z} \\
    \mathbf{L}_{k'} &:=& l_i\quad|\quad k'= \argmax_{k\in K}\,p(k | x_i),
\end{eqnarray}

where $\mathbf{L}_k$ denotes the \textit{last known location}, which is sequentially updated during inference time (the observed images need to be sorted by their timestamps). At the beginning of the inference, the \textit{last known location} $\mathbf{L}_k$ is initialized by setting it to the home location $H_k$, same as in the previous section.
Identically as in the previous section, we present two versions of the prior -- $\text{ML}_{\text{BG}}$ and $\text{ML}_\theta$ exploiting image background and image metadata respectively to determine camera geographical location.  \vspace{-0.1cm}\\

\paragraph{Time Decay (TD) Prior.} We also hypothesize that the time when the image is captured affects the class (identity) prior $p(k)$.  The prior $p(k)$ might change depending on the time of the day, month, or the season in general, but in our work, we opt to use the simplest time-based prior exploiting the fact that if we observed a given individual, we are likely going to observe it again; on the other hand, if we had not observed an individual for some time, the probability that we are going to observe it again decays with time since the last observation. 
More formally, we introduce the \textit{Time Decay Prior} as
\begin{eqnarray}
    \Psi_k(\tau_k, t_i) &= & \frac{\exp\big(-\beta\,|\tau_k - t_i| \big)}{Z} \\
    \tau_{k'} &:=& t_i\quad|\quad k'= \argmax_{k\in K}\,p(k | x_i),
\end{eqnarray}

where $\tau_k$ is the time when the identity $k$ was last seen, $t_i$ is the time when the image $x_i$ was taken, and $\beta$ is a decay parameter whose value is empirically found on the val. set. Again, we assume data are sorted by the acquisition date.

\section{Experiments}\label{sec:experiments}
\vspace{-0.1cm}


In experiments, we use two challenging 
datasets that have been collected over two decades, producing novel data with interesting properties, including timestamps and locations.
Both datasets feature species that are often visually indistinguishable to non-experts, exhibit changing appearance over time, and have a long-tailed distribution. They include diverse poses and acquisition conditions.
The data have not been seen, with high probability%
\footnote{
    Access to the dataset requires signing a Non-Disclosure Agreement (NDA) with {\it Friends of the Earth, Czech Republic} due to animal safety concerns.
},%
by any LLM. Thus, the test set presents new data even for methods that exploit LLMs for re-identification.\\

\paragraph{Data preprocessing:} For both datasets, we first detect animals with MegaDetector~\cite{beery2019efficient}, and then we feed forward the detected bounding box with the image into the Segment Anything Model~\cite{kirillov2023segany} to separate foreground and background pixels. The resulting masks are cropped and resized to $256\times256$ resolution.

We use the EfficientNet-B3 model~\cite{tan2019efficientnet} pre-trained on ImageNet-1k~\cite{imagenet} and train it for 100 epochs using AdamW optimizer~\cite{adamw}. The initial LR is set to 0.001, and it is then gradually lowered using a cosine annealing. While training, we use random crop, random flip, and random color jitter data augmentations. The $\lambda$ weighting parameter in the \Cref{eq:finalloss} was set to $0.1$ in all experiments and ablations, but we found that varying the parameter value has minimal impact. 

\subsection{Eurasian lynx} 
\begin{figure}[t]
    \scriptsize
    \centering
    \setlength\tabcolsep{1pt}
    \begin{tabular}{ccc@{\hspace{5pt}}|@{\hspace{5pt}}ccc}
        \multicolumn{3}{c|@{\hspace{5pt}}}{\small \hspace{10pt}\textbf{\scriptsize Training set example}} & \multicolumn{3}{c}{\scriptsize \textbf{Test set predictions}} \\
        \multicolumn{3}{c|@{\hspace{5pt}}}{\hspace{10pt}\textbf{data until 2019}} & \multicolumn{3}{c}{\textbf{data after 2020}} \\
        & \rotatebox[origin=c]{90}{\textbf{Albina}} &
        \includegraphics[height=2.2cm, width=3.2cm, align=c, trim=0 0 40 0, clip]{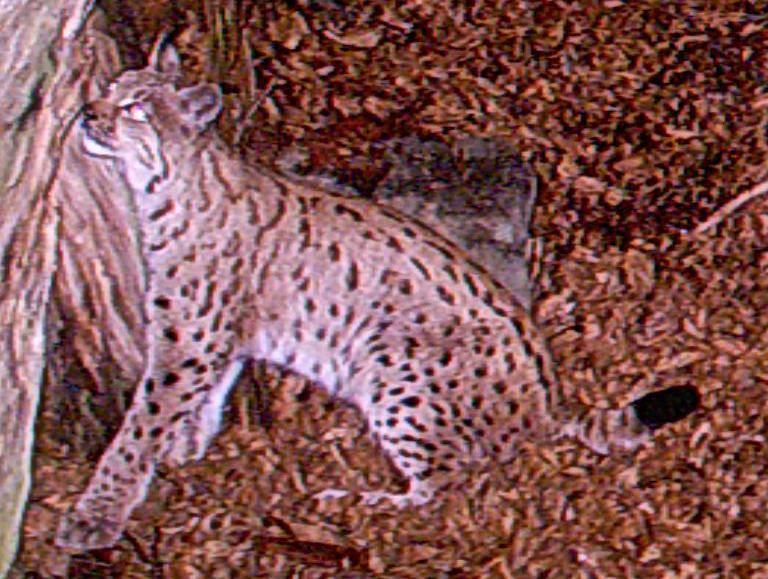} 
        & \includegraphics[height=2.2cm, width=3.3cm, align=c]{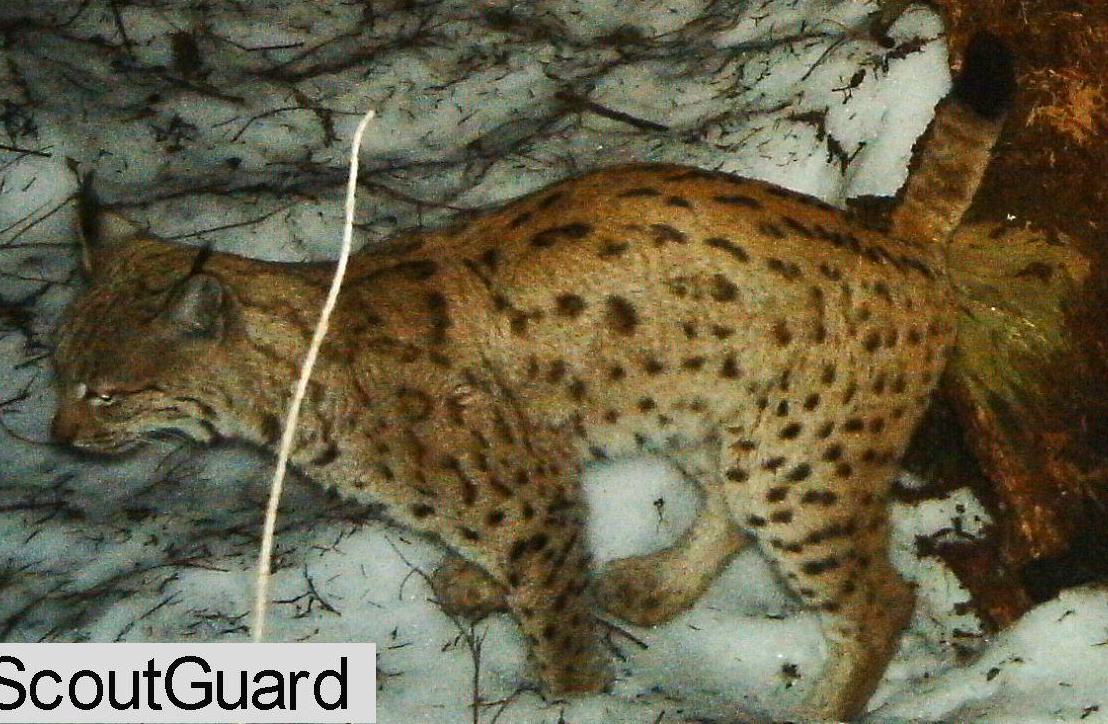} 
        & \includegraphics[height=2.2cm, width=1.2cm, align=c, trim=10 0 200 0, clip]{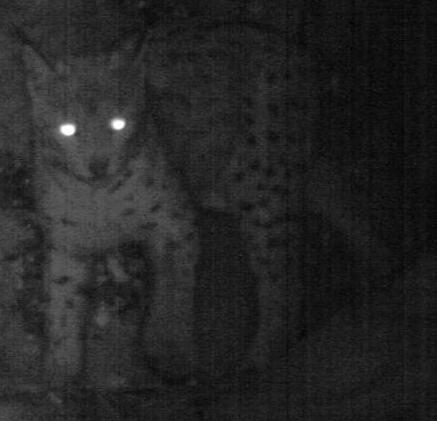} 
        & \includegraphics[height=2.2cm, width=2.8cm, align=c]{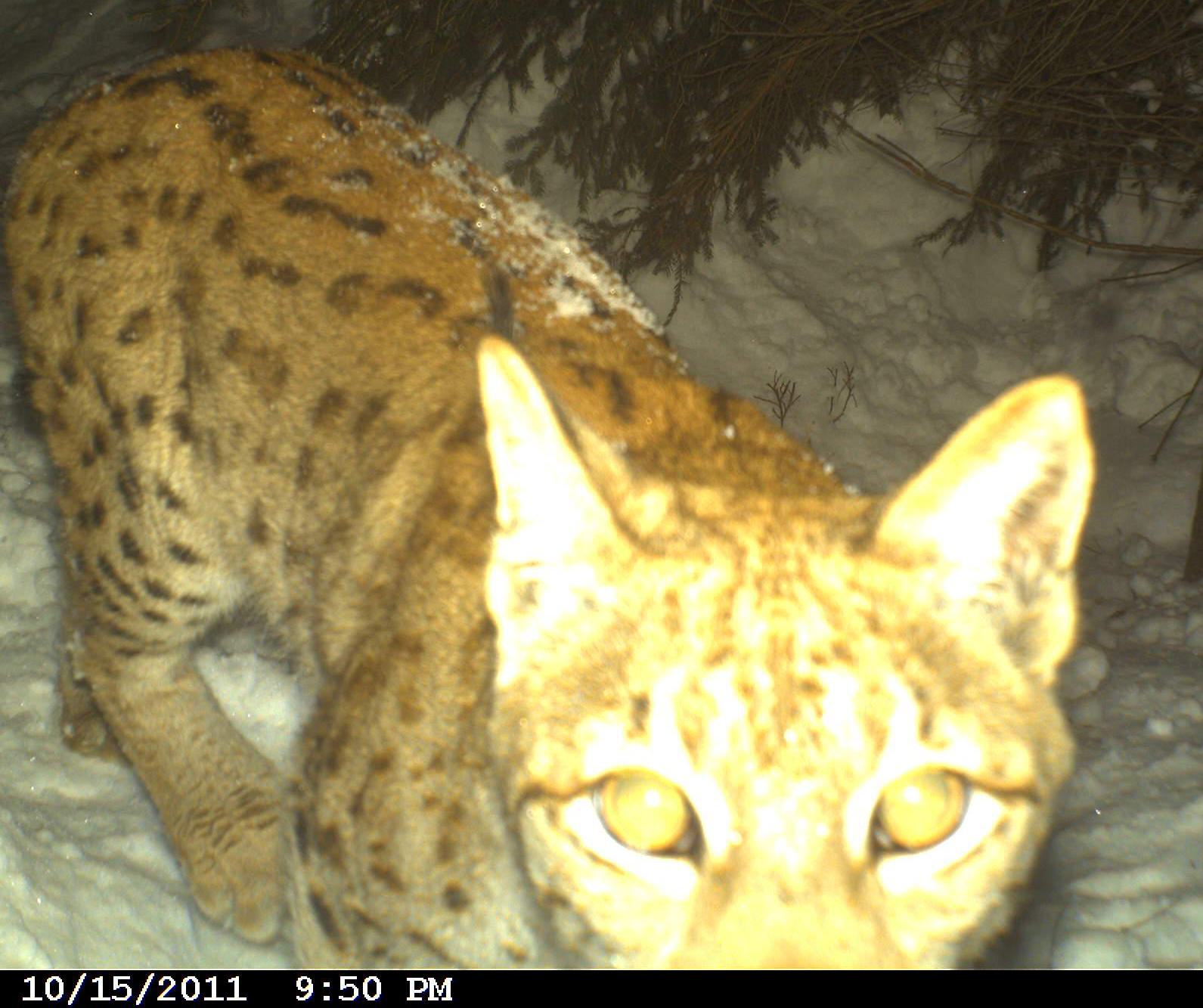} 
        \\
         && & \textit{Albina}  \color{darkgreen} \cmark & \textit{Albina}  \color{darkgreen} \cmark & \textit{Albina}  \color{darkgreen} \cmark \\ [2pt]
        \rotatebox[origin=c]{90}{\scriptsize\textbf{Ground truth}}\hspace{0.5em} &  \rotatebox[origin=c]{90}{\textbf{Ji\v{r}\'{i}}} &
        \includegraphics[height=2.2cm, width=3.2cm, align=c, trim=0 200 0 200, clip]{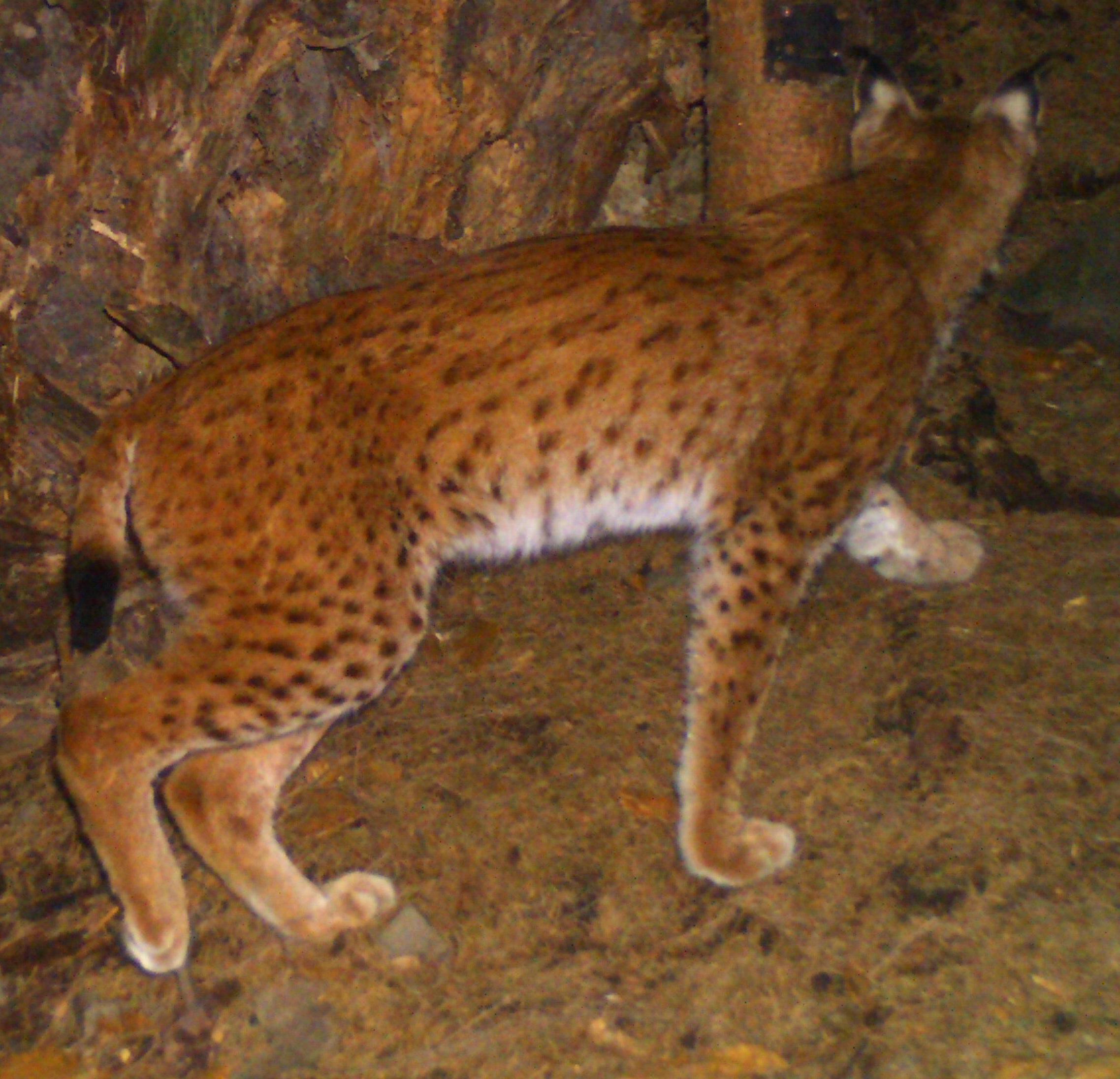} 
        & \includegraphics[height=2.2cm, width=3.3cm, align=c, trim=20 0 20 0, clip]{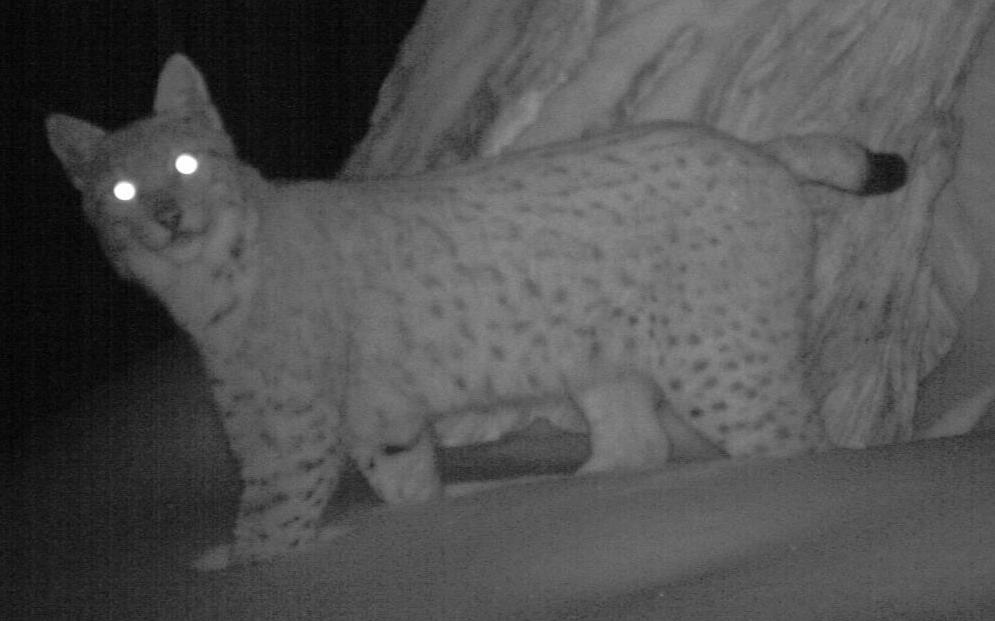} 
        & \includegraphics[height=2.2cm, width=1.2cm, align=c]{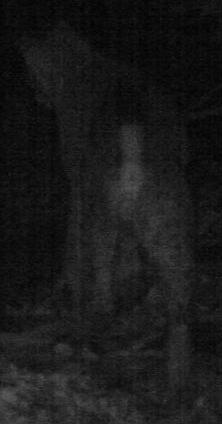}         
        & \includegraphics[height=2.2cm, width=2.8cm, align=c]{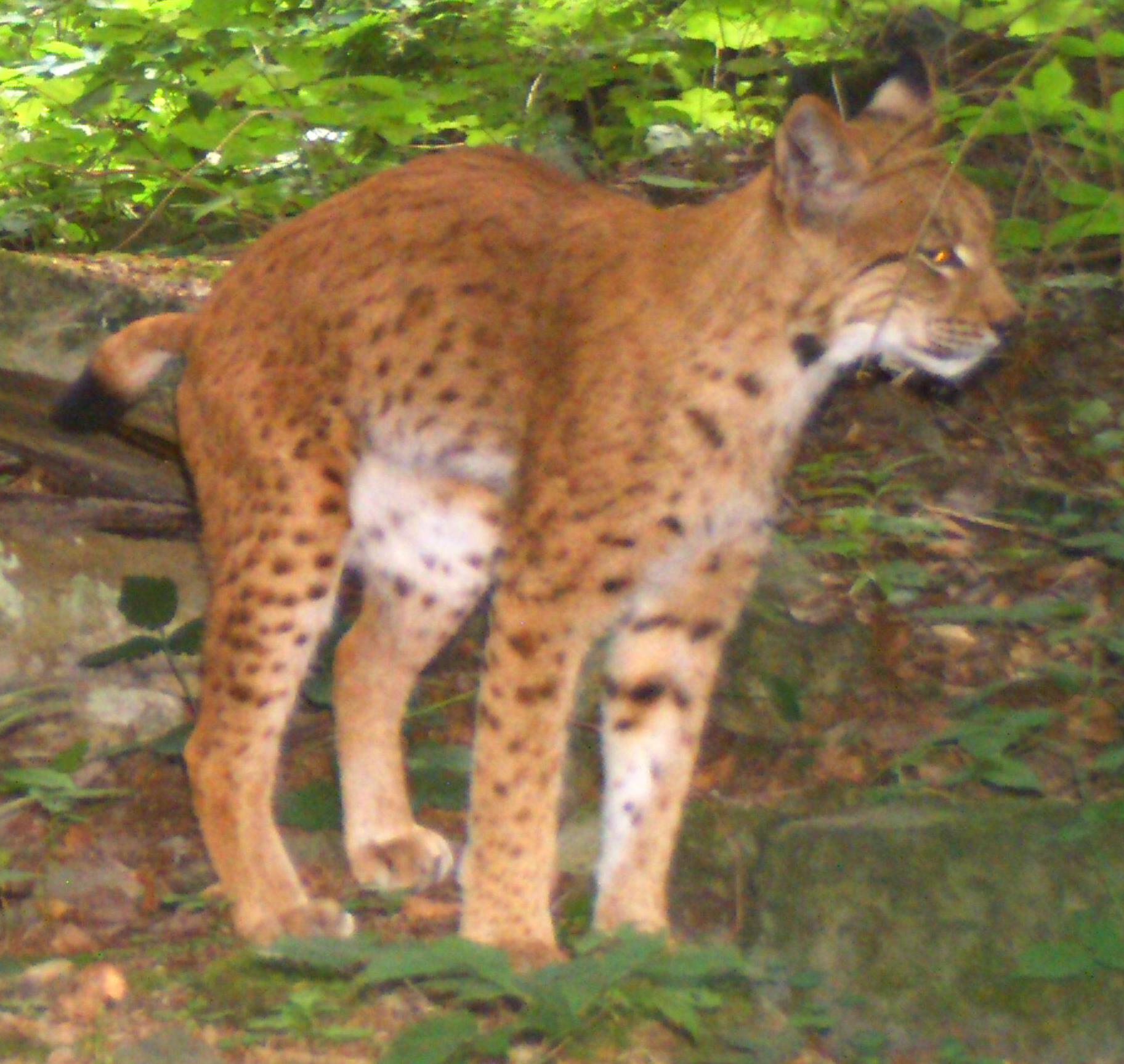} \\
        && & \textit{Ji\v{r}\'{i}}  \color{darkgreen} \cmark & \textit{Ji\v{r}\'{i}}  \color{darkgreen} \cmark & \textit{Ji\v{r}\'{i}}  \color{darkgreen} \cmark  \\ [2pt]
        \rotatebox[origin=c]{90}{\scriptsize\textbf{Ground truth}}\hspace{0.5em} & \rotatebox[origin=c]{90}{\textbf{Rufus}} &
        \includegraphics[height=2.2cm, width=3.2cm, align=c]{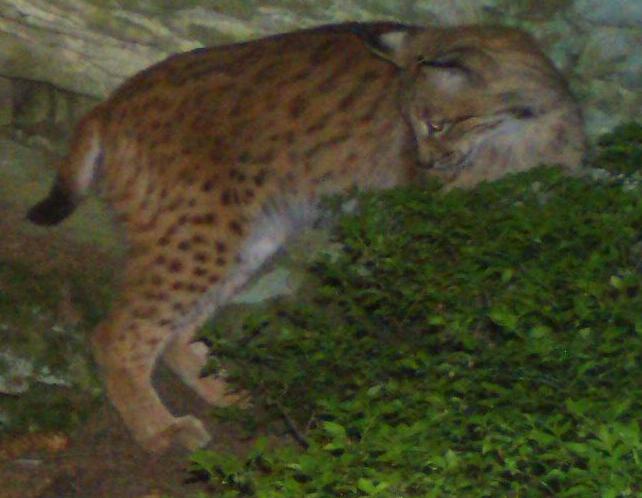} 
        & \includegraphics[height=2.2cm, width=3.3cm, align=c]{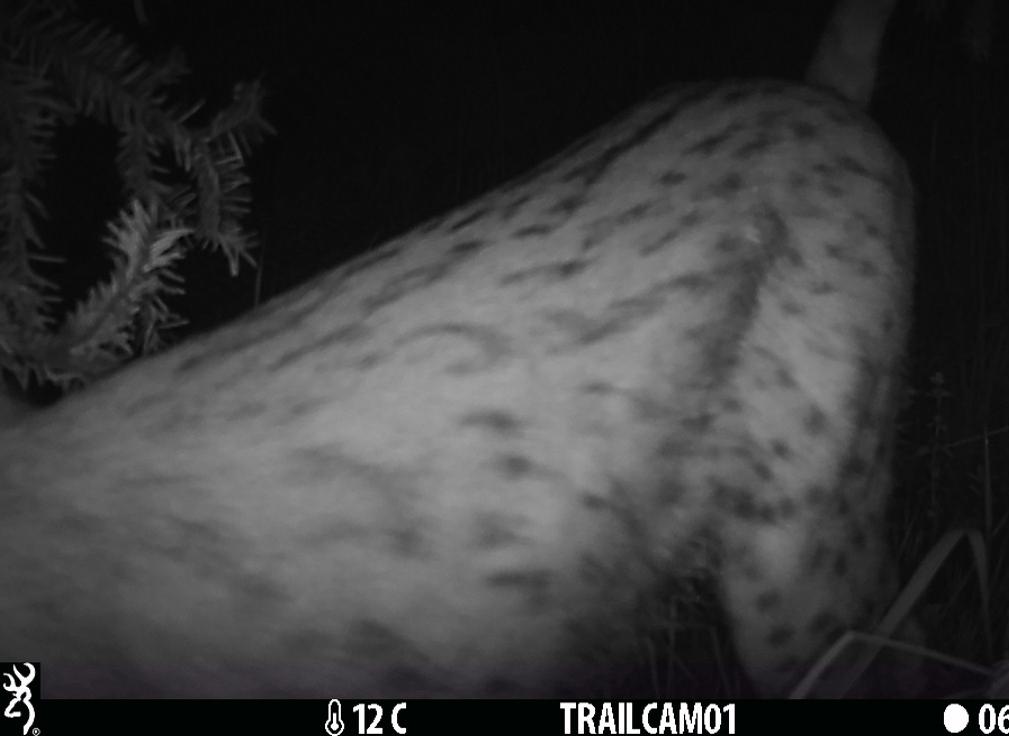}
        & \includegraphics[height=2.2cm, width=1.2cm, align=c]{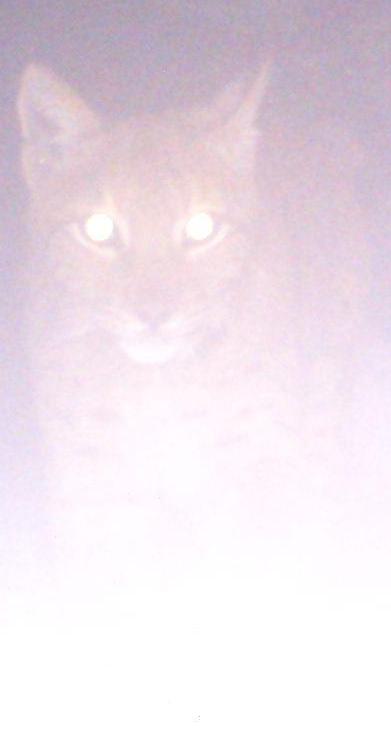} 
        & \includegraphics[height=2.2cm, width=2.8cm, align=c]{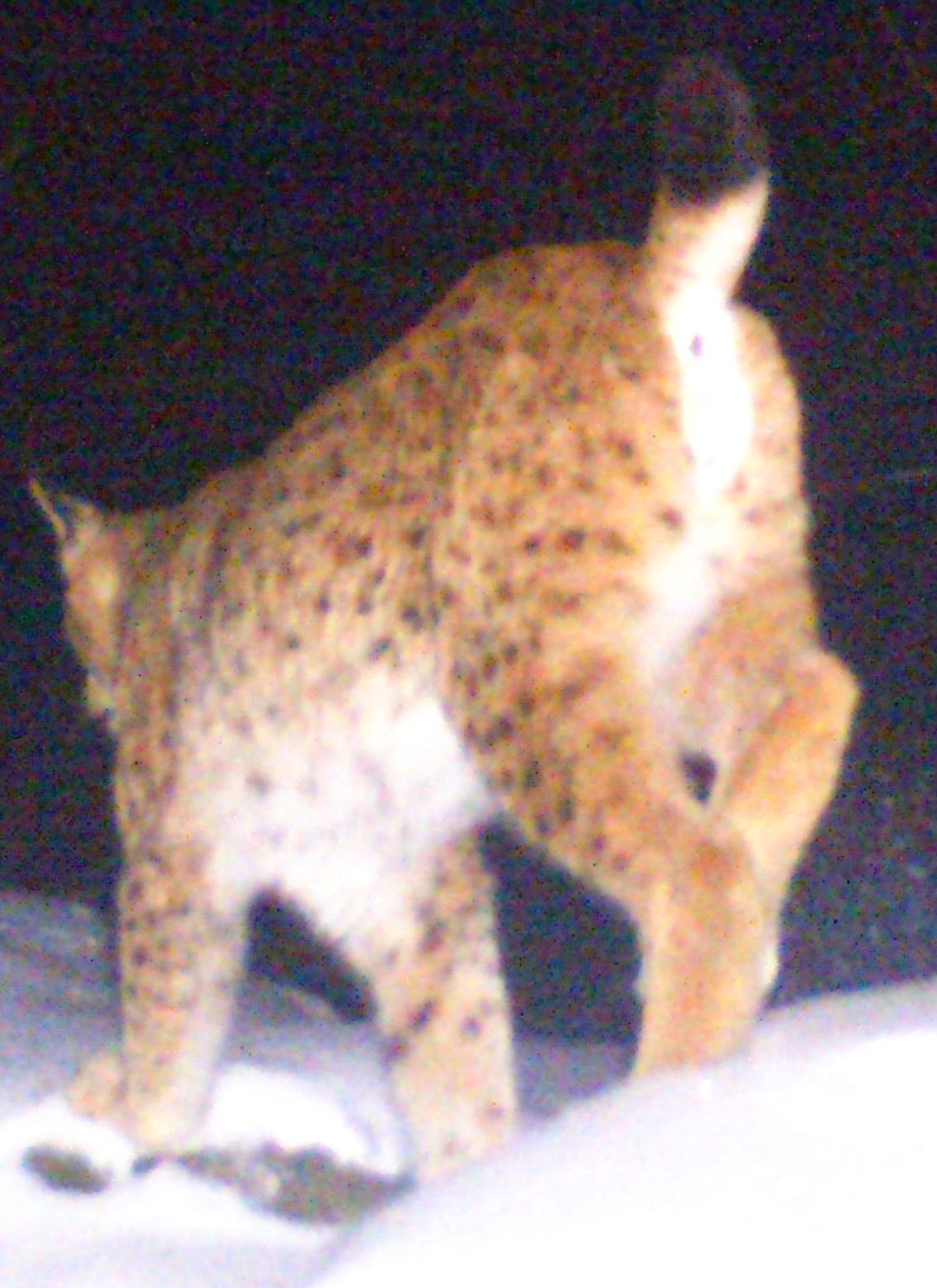}
         \\ 
        && & \textit{Kr\'{a}l}  \color{darkred} \xmark & \textit{Rufus}  \color{darkgreen} \cmark & \textit{Rufus}  \color{darkgreen} \cmark  \\
    \end{tabular}
    \caption{Examples of lynx re-identification.
    }
    \label{fig:lynx-qualitative}
    \vspace{-10pt}
\end{figure}

The dataset originates from long-term monitoring in the West Carpathian mountains using camera traps comprising 6,743 images of 103 individuals over the past 15 years at 81 locations
It exhibits diverse acquisition conditions and significant variability in individual appearances, presenting a complex identification challenge, see~\Cref{fig:lynx-qualitative}. Images captured before  2020 form the \textit{training set}, newer images the \textit{test set}. 
Such temporal split is customary in long-term re-identification research~\cite{Dula2021} and aligns with wildlife conservation efforts, where timely processing of new data is crucial for understanding migration patterns and identifying new individuals.
For this dataset, we employ the Home and Migrating Location priors, which are natural models for wild cat behavior. Through our experiments and variations, we maintain hyper-parameters $\alpha = 2.5$.

\paragraph{Results.} 

Compared to the baseline method, which uses a cropped animal, masking out background pixels slightly worsens overall accuracy. In contrast, at the same time, accuracy at new locations\footnote{We define a \emph{new location} as a location in the test set, where an individual was \textit{not seen in the training set}. This is a \textit{crucial metric} in animal re-identification as timely information about animals migrating to new locations is key to nature preservation.} increases (see \Cref{tab:lynx}). This is expected because the network is now forced to base its decision only on the individual's appearance, which is mostly independent of the location where the individual is captured. The problem with location embedded in individuals' identification is further demonstrated by the \emph{Background} method in \Cref{tab:lynx}, which only accepts background pixels as the input. Yet, it achieves competitive results in recognizing individuals in known locations (but fails completely for new locations). Replacing the traditional CE loss with our PITS loss (see \Cref{sec:pitsloss}) yields a small yet significant improvement, which we believe is due to the fact that the network is no longer forced to precisely recognize identities with a small number of training samples and therefore can ``give up'' on certain samples by increasing their per-instance temperature 
$T_i$ \Cref{eq:perinstancetemperaturescaling}, focusing more on the less ambiguous samples.
The biggest gain in accuracy, however, comes from incorporating the instance prior 
$\Psi_k$
to the network calibrated using the PITS loss, where for the Migrating Location (ML) prior, the \textbf{accuracy is increased by more than 11 percentage points} over the baseline method. The accuracy at new locations is increased by 7 percentage points over the baseline. However, the absolute accuracy of 
19.6\% clearly shows this still remains an open problem.

\begin{table}[t]
  \setlength{\tabcolsep}{0.85em}
  \centering
   \caption{Eurasian lynx re-identification accuracy. The \emph{new location} column denotes accuracy at locations where a given individual was not observed during training.}
  \begin{tabular}{lcc|cc}
    \toprule
          &      & Background      & \multicolumn{2}{c}{Accuracy} \\
    Input & Loss & model $\Psi_k$ &  \textit{overall} & \textit{new location} \\
    \midrule
    Whole image \textit{(baseline)} & CE &  -- &  49.7\% & 13.9\%\\
    Background & CE & -- & 27.1\% & ~\,3.5\% \\ 
    Foreground & CE &-- &49.3\% & 16.3\% \\ 
    Foreground & PITS & -- & 52.5\% & 19.1\% \\ 
    \midrule
    Foreground + $\Psi_k$ & PITS & $\text{HL}_{\text{BG}}$ & 56.5\% & 15.3\% \\
    Foreground + $\Psi_k$ & PITS & $\text{HL}_\theta$ &\underline{59.7}\% & \underline{17.7}\% \\    
    Foreground + $\Psi_k$ & PITS & $\text{ML}_{\text{BG}}$ & 57.1\% & 17.2\% \\ 
    Foreground + $\Psi_k$ & PITS & $\text{ML}_\theta$ & \textbf{60.9\%} & \textbf{19.6\%} \\ 
    \bottomrule
  \end{tabular}
     \vspace{-5pt}
  \label{tab:lynx}
\end{table}

\subsection{Loggerhead sea turtle}

\begin{figure}[h]
    \vspace{-10pt}
    \scriptsize
    \centering
    \setlength\tabcolsep{1pt}
    \begin{tabular}{ccc@{\hspace{5pt}}|@{\hspace{5pt}}ccc}
        \multicolumn{3}{c|@{\hspace{5pt}}}{\small \hspace{10pt}\textbf{\scriptsize Training set example}} & \multicolumn{3}{c}{\scriptsize \textbf{Test set predictions}} \\
        \multicolumn{3}{c|@{\hspace{5pt}}}{\hspace{10pt}\textbf{(data until 2019)}} & \multicolumn{3}{c}{\textbf{(data after 2020)}} \\
        
        & \rotatebox[origin=c]{90}{\textbf{t028}} &
        \includegraphics[height=2.0cm, align=c]{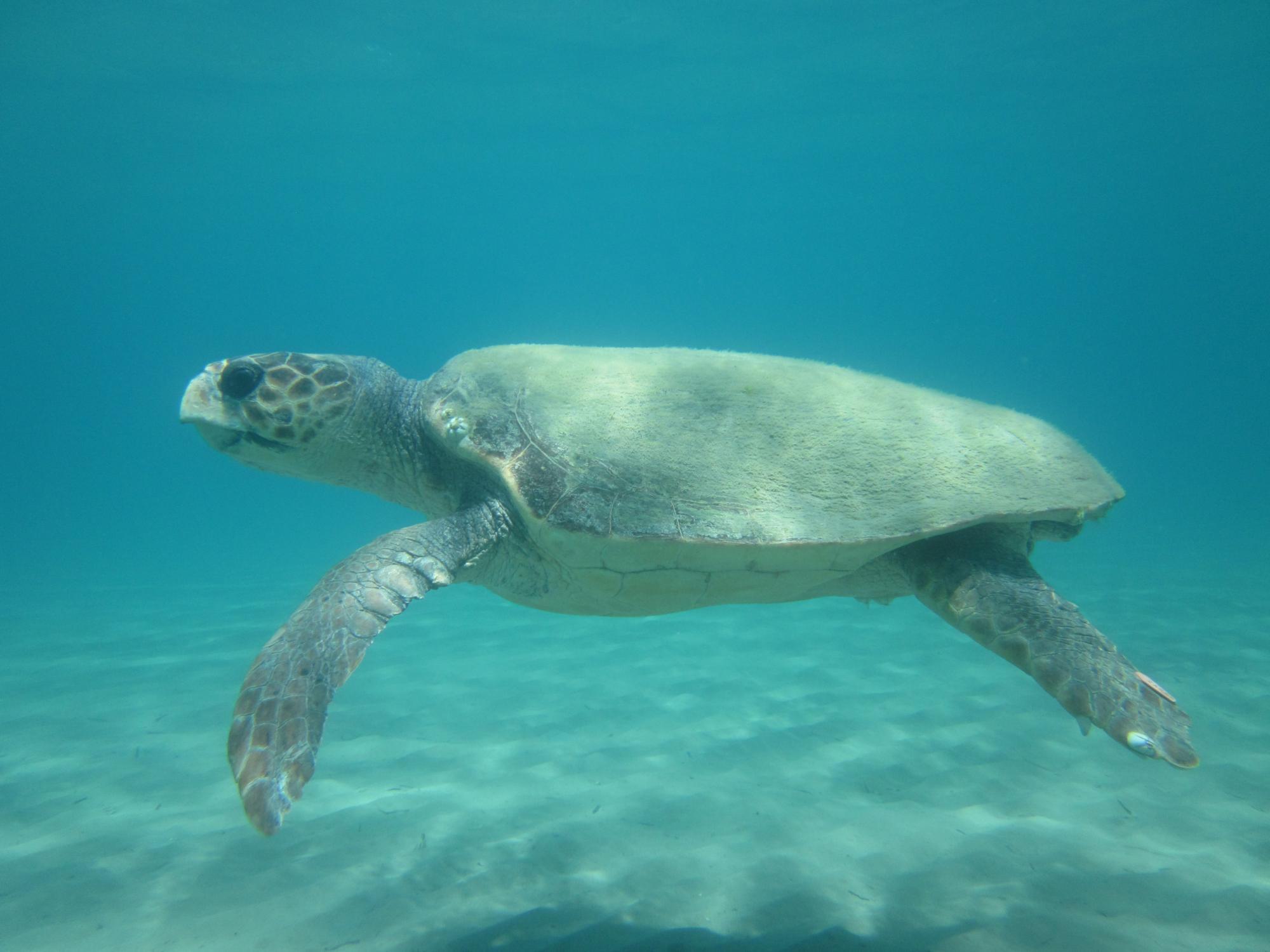} 
        & \includegraphics[height=2.0cm, align=c, trim=80 0 110 0, clip]{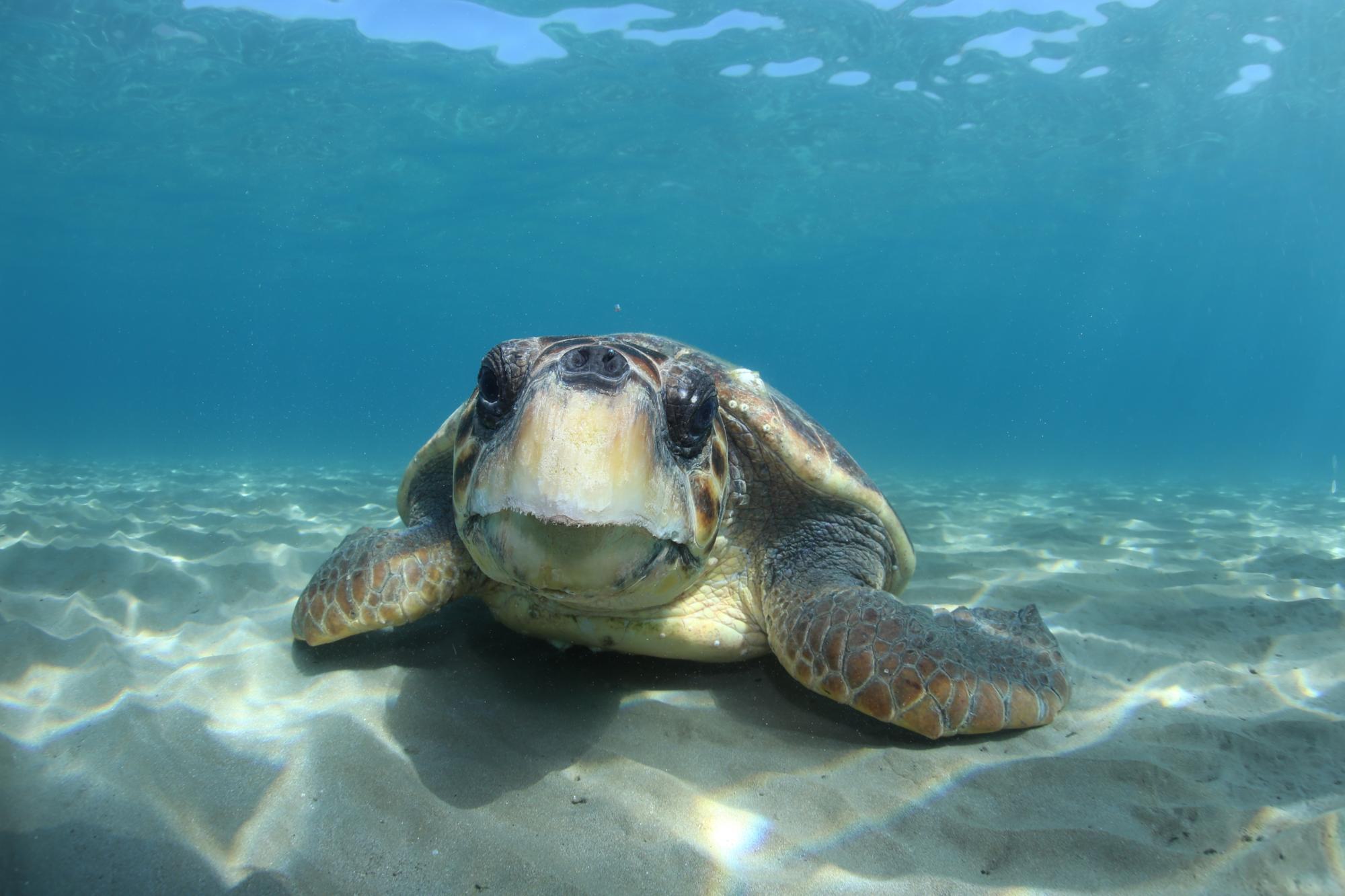} 
        & \includegraphics[height=2.0cm, align=c, trim=230 0 0 0, clip]{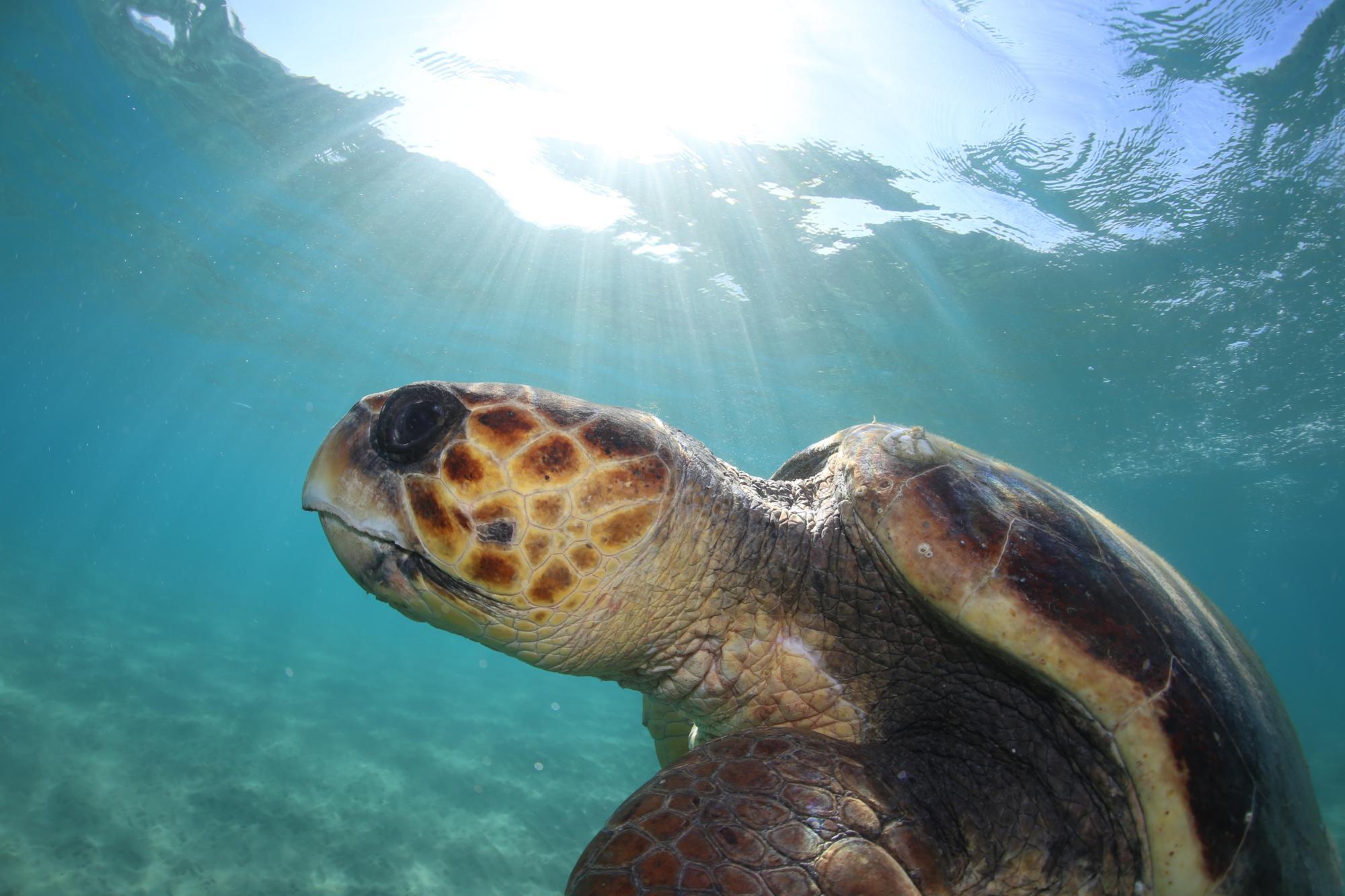} 
        & \includegraphics[height=2.0cm, align=c, trim=100 0 130 0, clip]{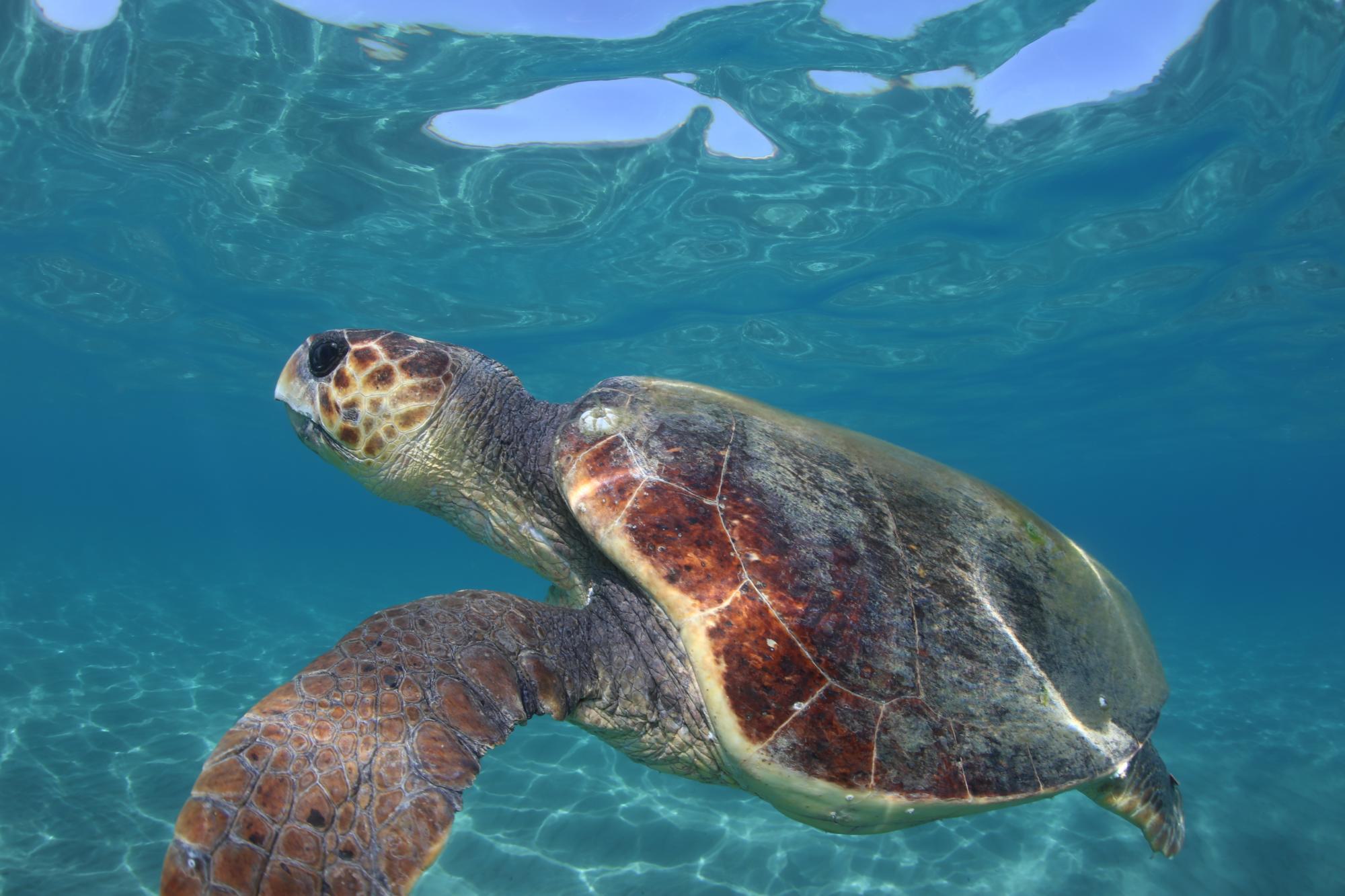} 
        \\[3pt]
         && & t028  \color{darkgreen} \cmark & t028  \color{darkgreen} \cmark & t110  \color{darkred} \xmark  \\  [3pt]       
        \rotatebox[origin=c]{90}{\scriptsize{\textbf{Ground truth}}}\hspace{0.5em} & \rotatebox[origin=c]{90}{\textbf{t110}} &
        \includegraphics[height=2.0cm, align=c, trim=100 0 130 0, clip]{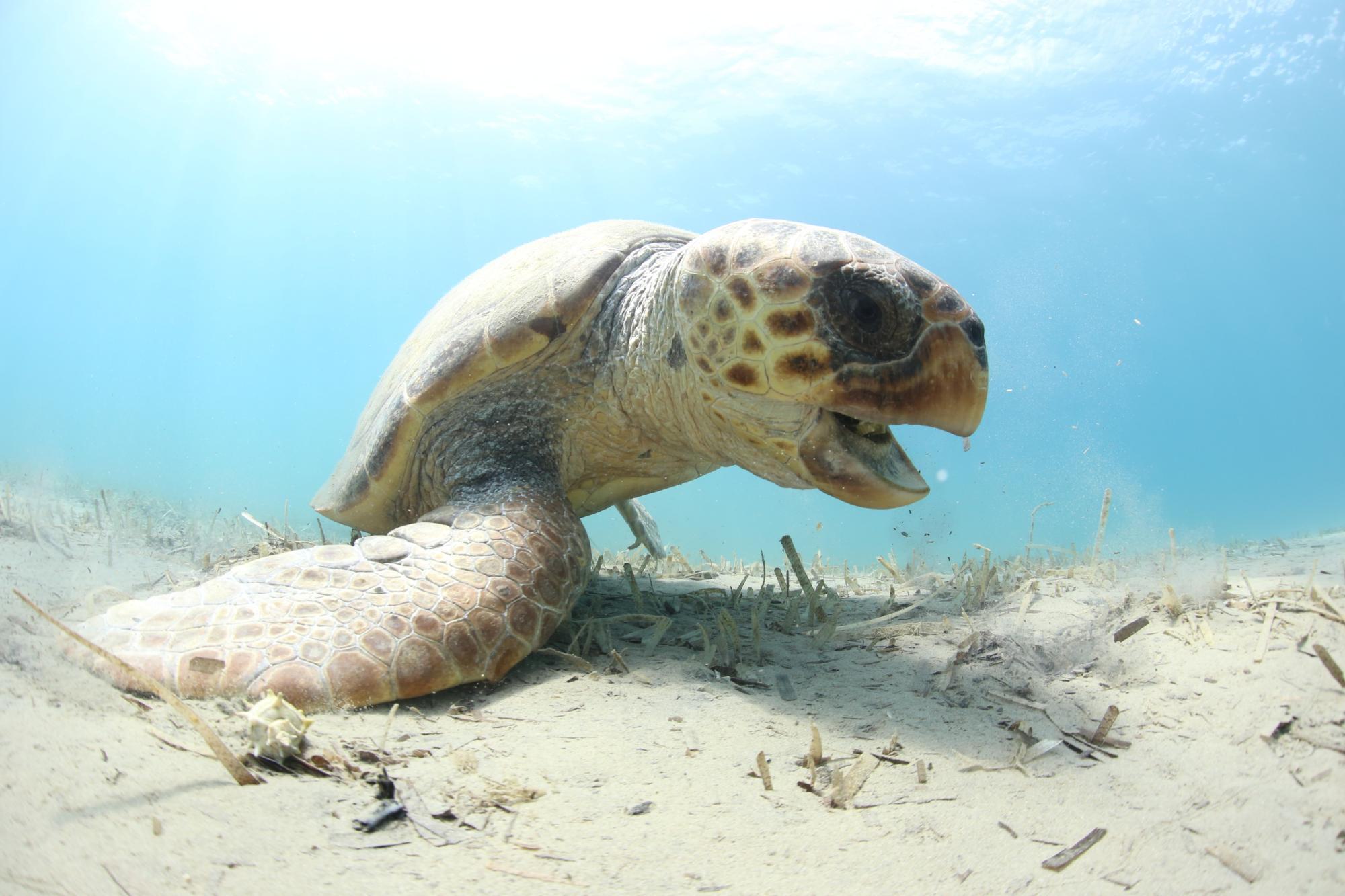} 
        & \includegraphics[height=2.0cm, align=c, trim=100 0 130 0, clip]{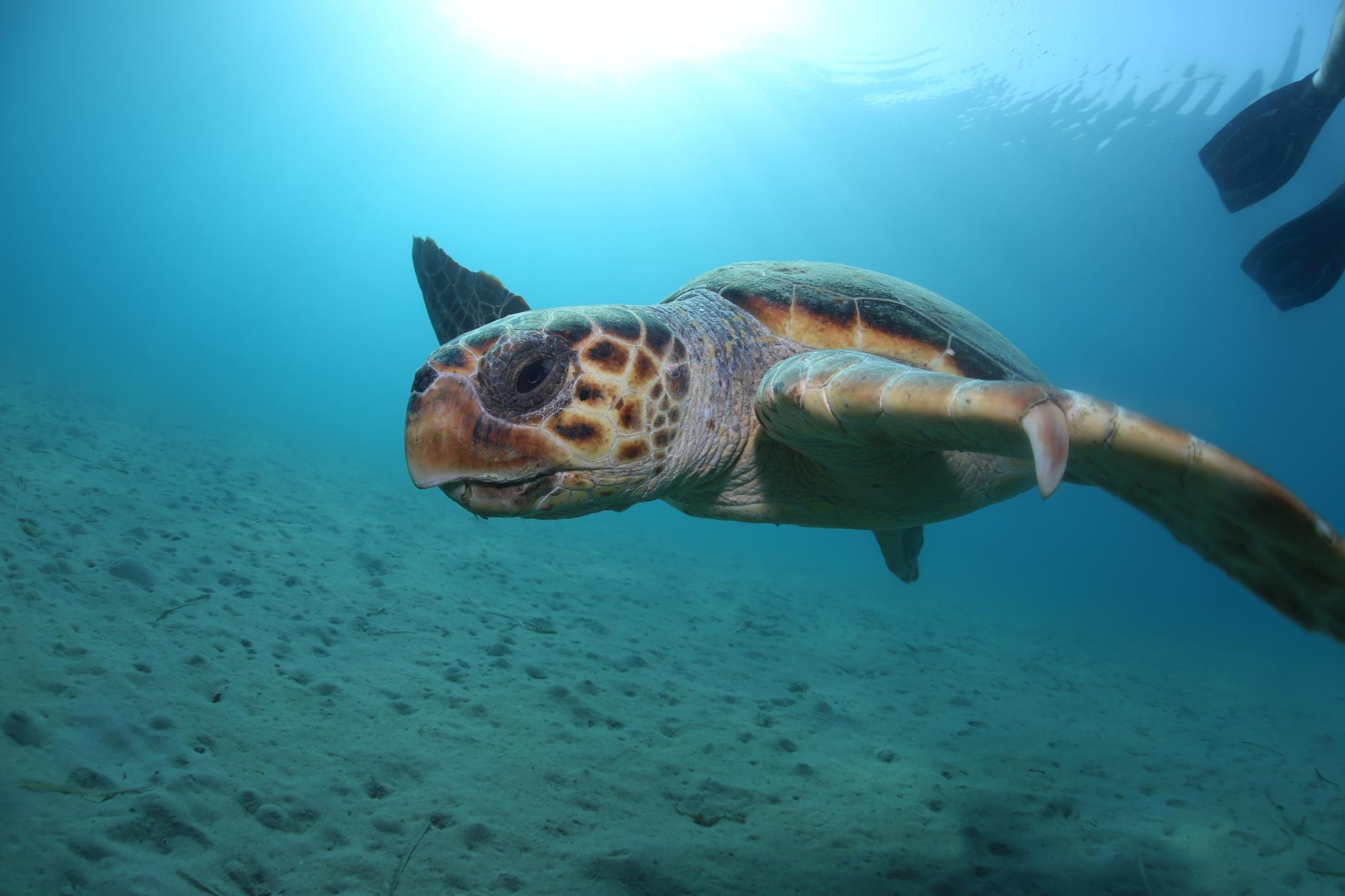} 
        & \includegraphics[height=2.0cm, align=c, trim=100 0 130 0, clip]{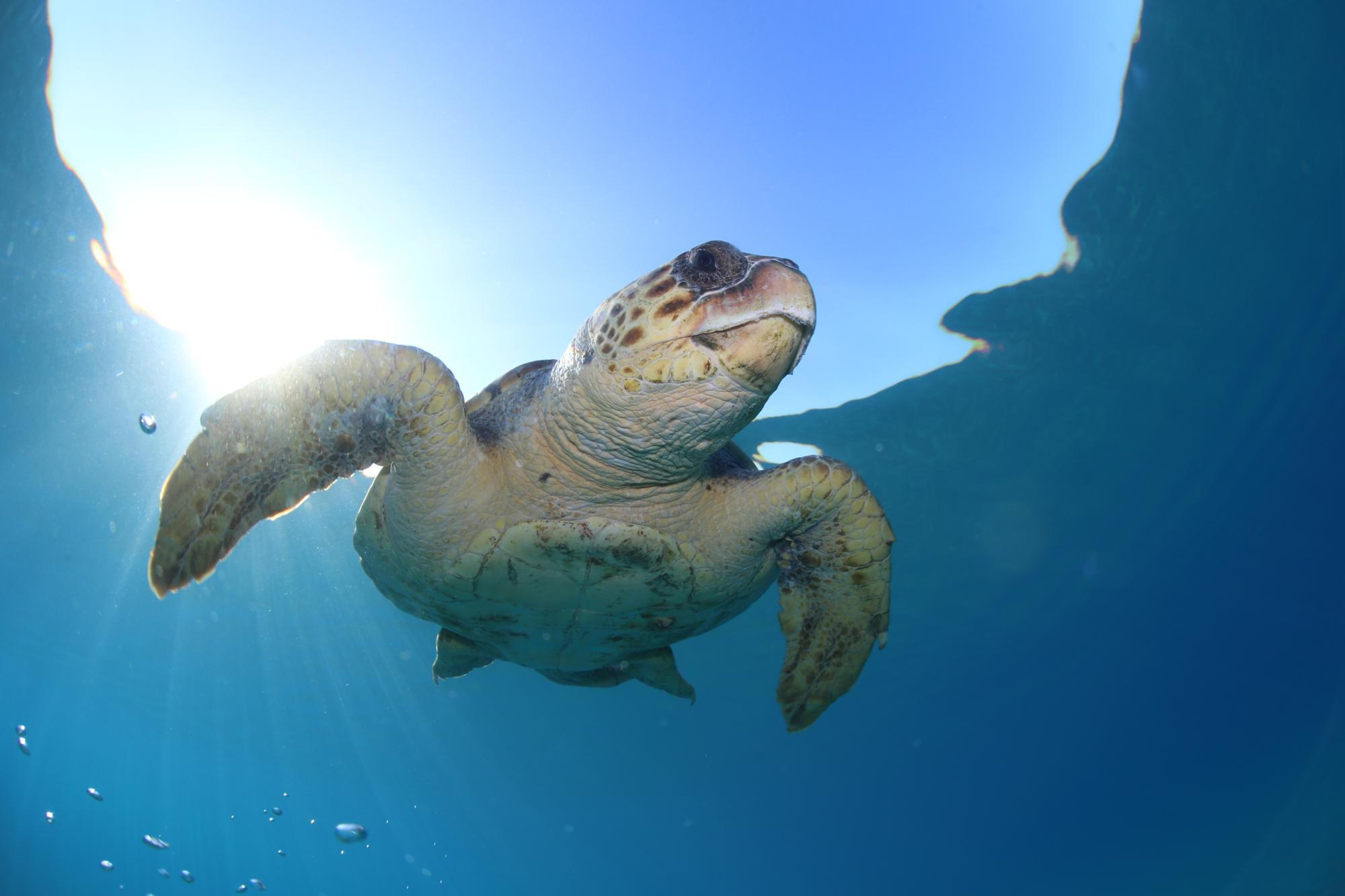} 
        & \includegraphics[height=2.0cm, align=c, trim=100 0 130 0, clip]{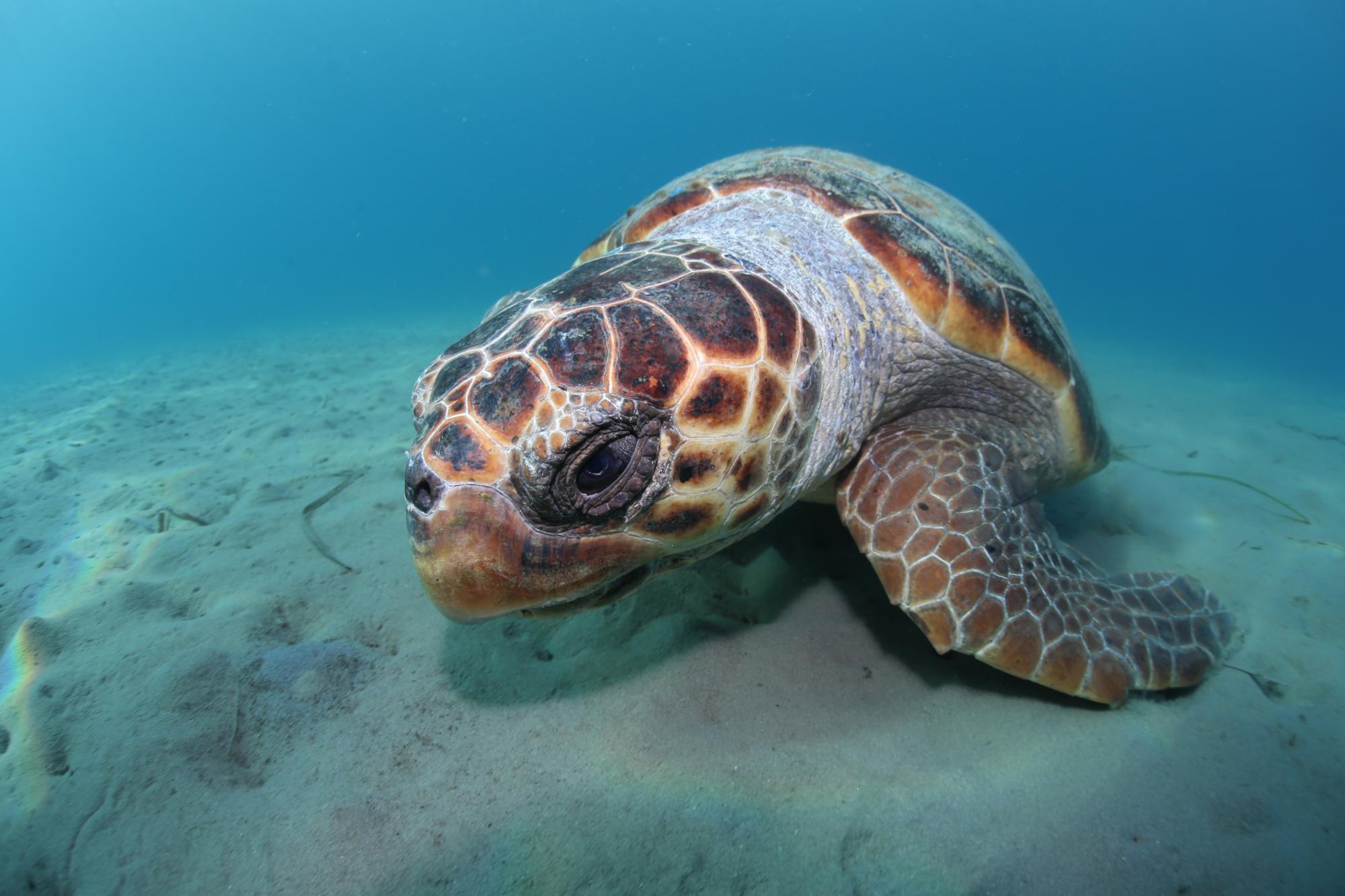} 
        \\       
        && & t110  \color{darkgreen} \cmark & t110  \color{darkgreen} \cmark & t028  \color{darkred} \xmark   \\ [3pt]
         & \rotatebox[origin=c]{90}{\textbf{t230}} &
        \includegraphics[height=2.0cm, align=c, trim=100 0 130 0, clip]{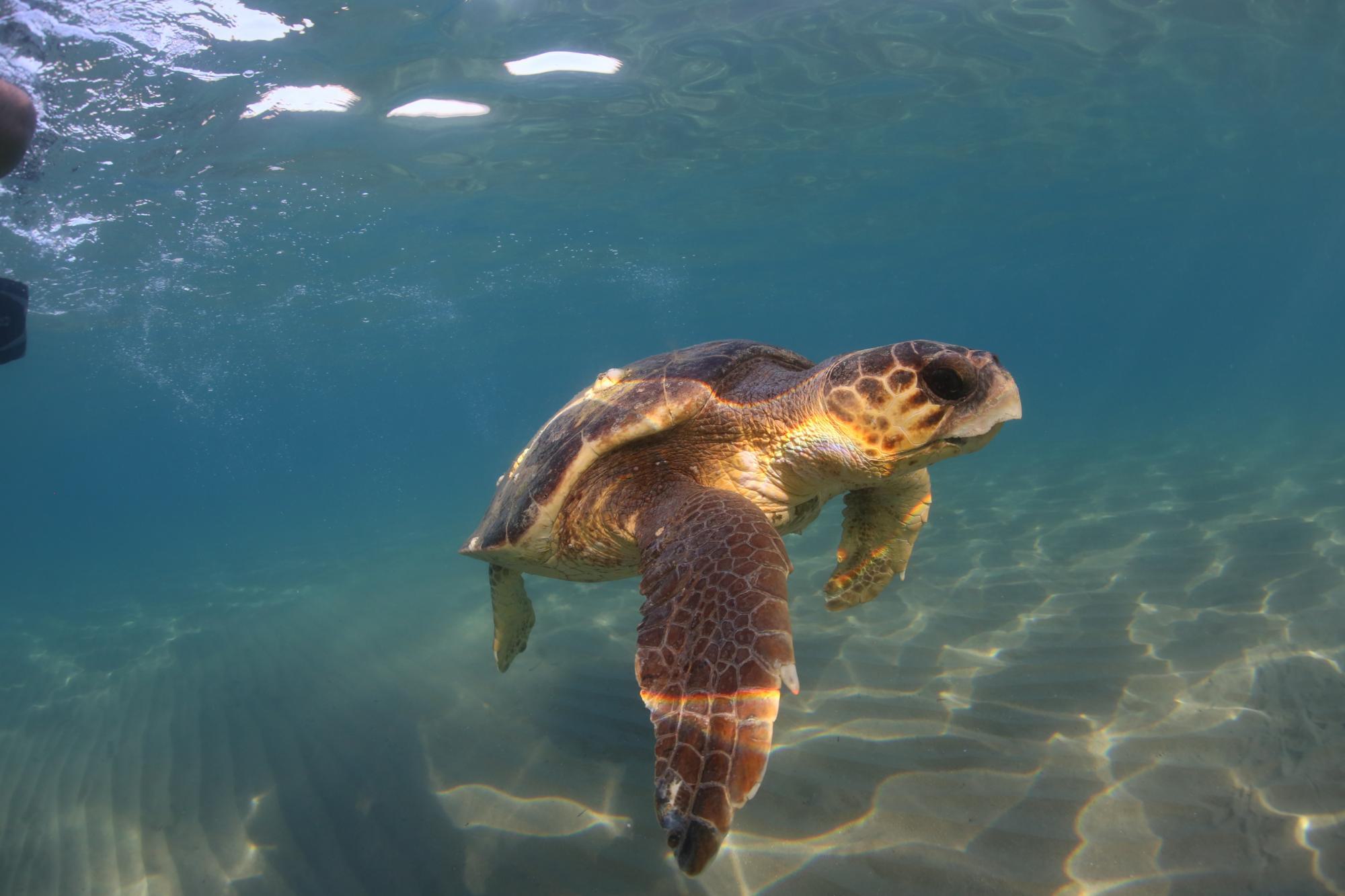} 
        & \includegraphics[height=2.0cm, align=c, trim=0 0 230 0, clip]{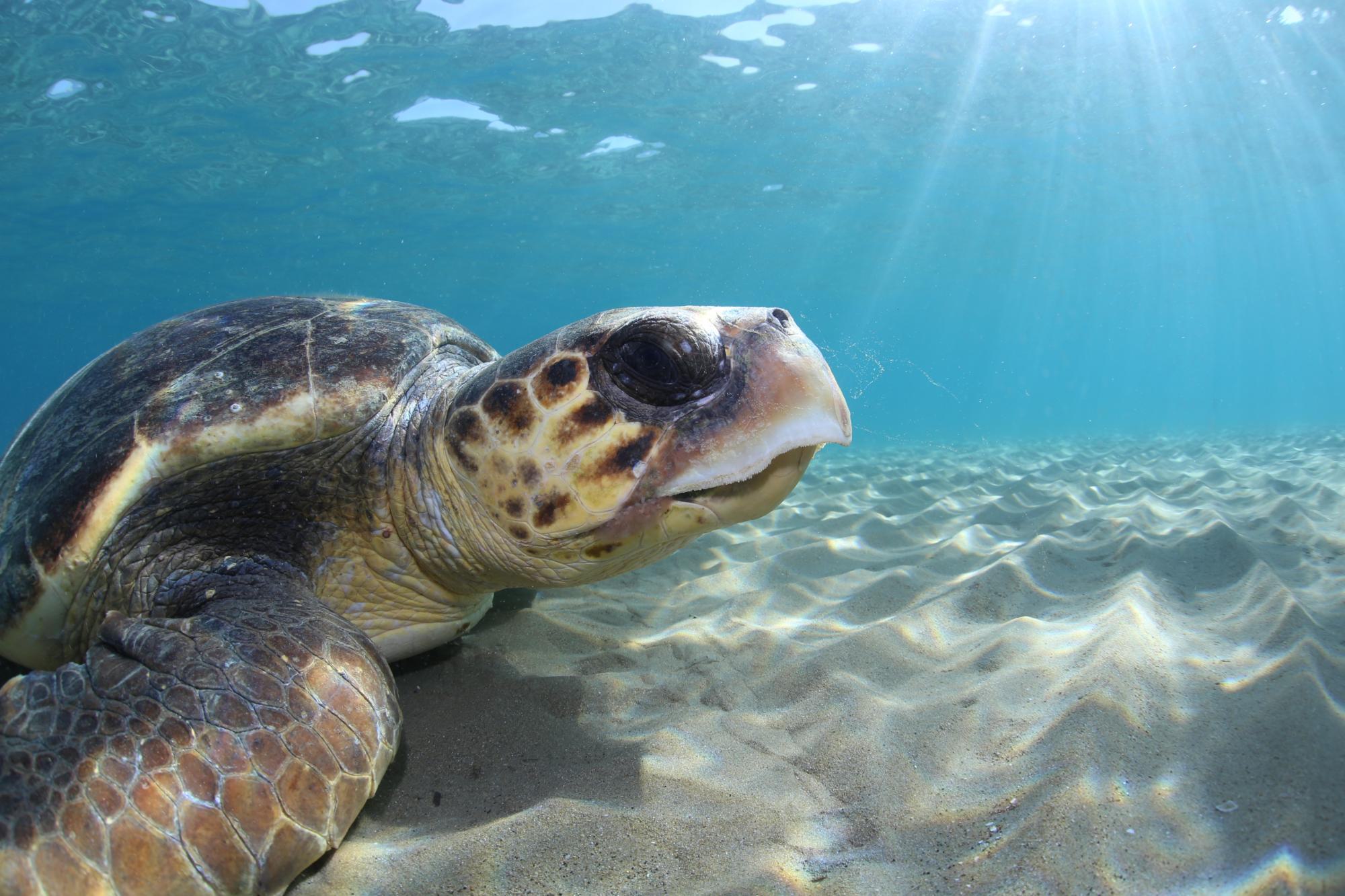} 
        & \includegraphics[height=2.0cm, align=c, trim=230 0 0 0, clip]{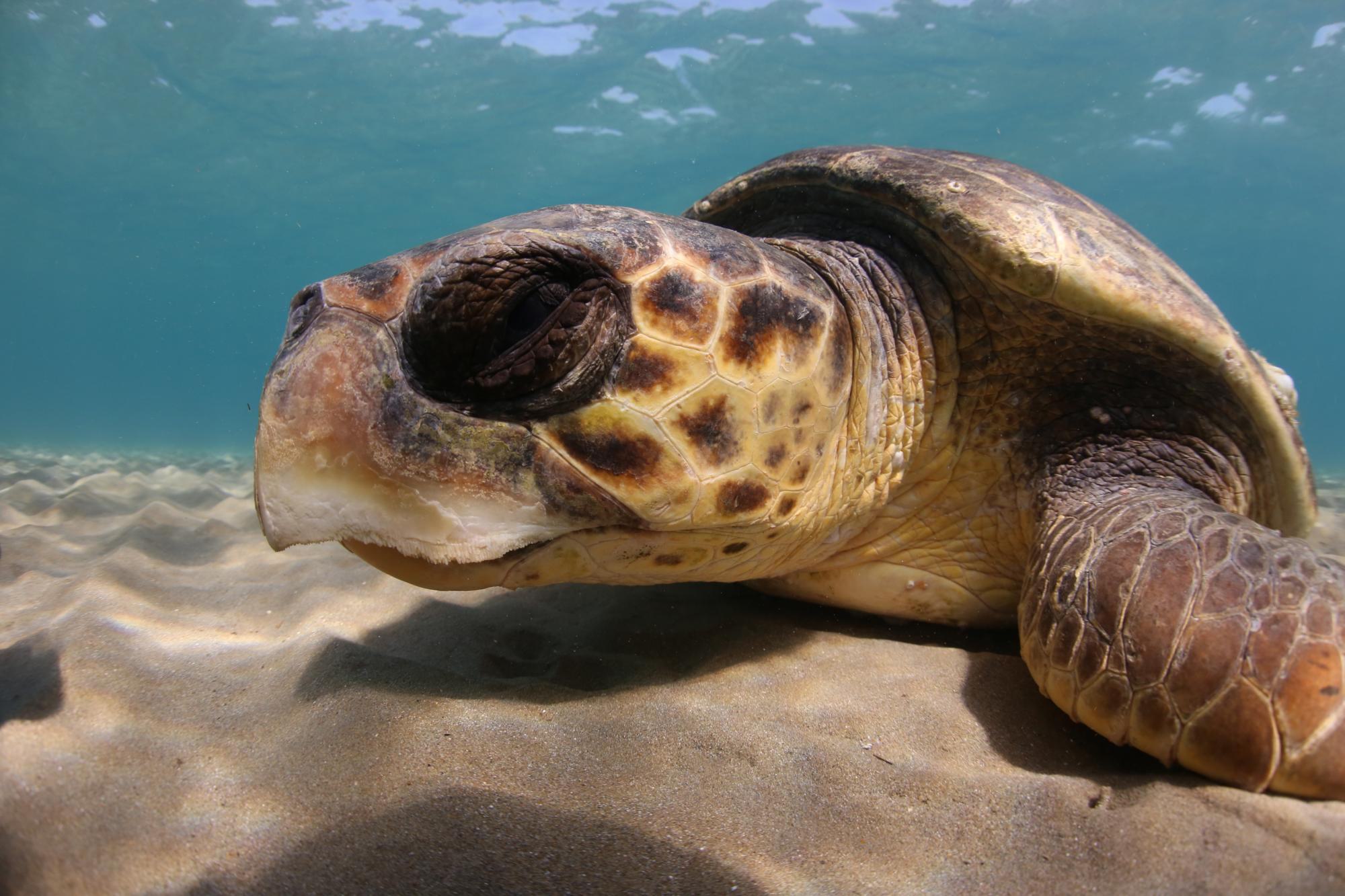} 
        & \includegraphics[height=2.0cm, align=c, trim=0 0 230 0, clip]{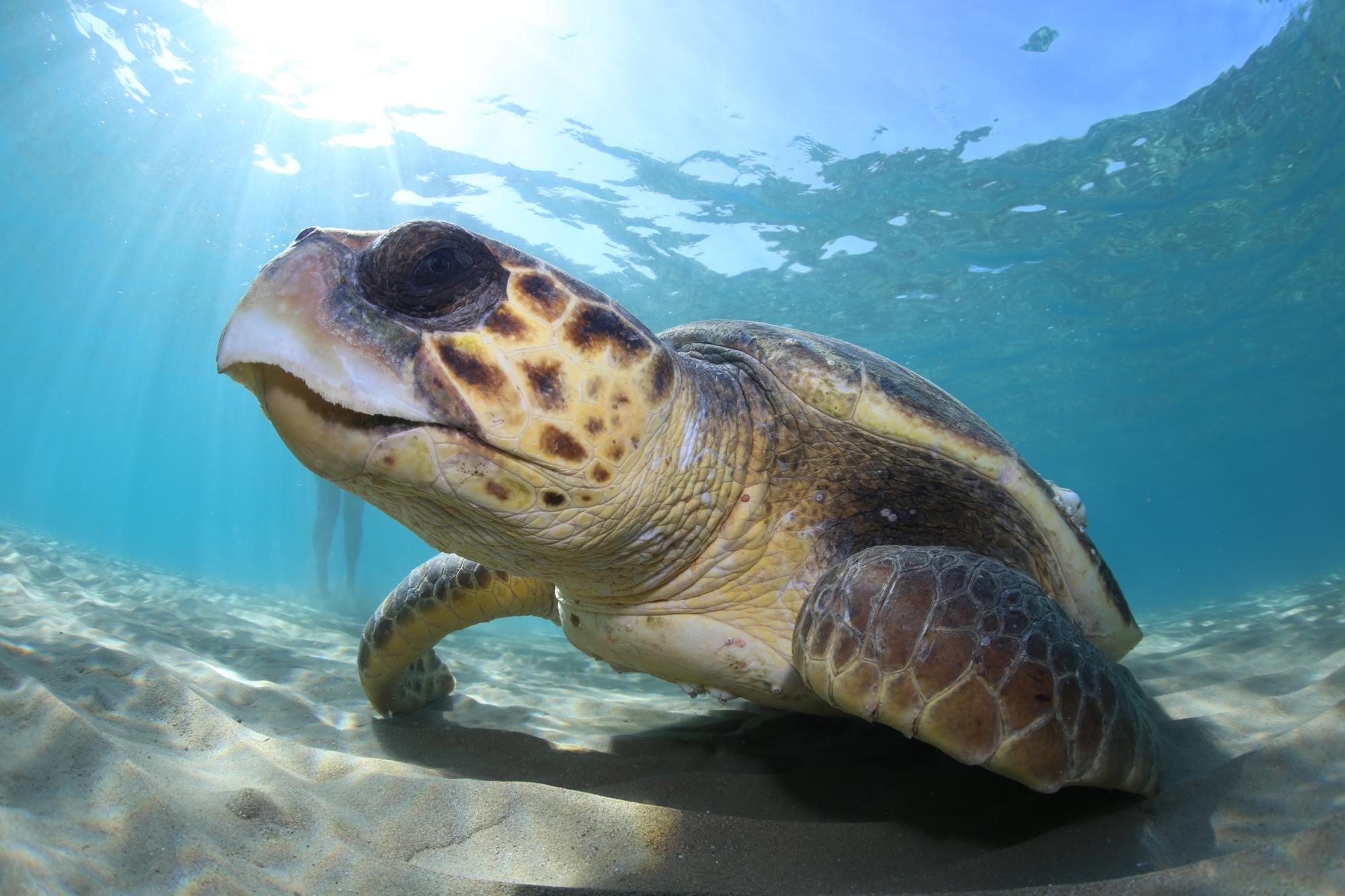} 
        \\
        && & t230  \color{darkgreen} \cmark & t230  \color{darkgreen} \cmark & t067  \color{darkred} \xmark   \\ 
    \end{tabular}
    \vspace{-5pt}
    \caption{Examples of Loggerhead sea turtle (\textit{Caretta caretta}) re-identification.
    }
    \label{fig:turtles-qualitative}
    \vspace{-10pt}
\end{figure}

The SeaTurtleID2022\,\cite{seaturtleid2022} dataset contains 8,729 photographs of 438 unique individuals.
Besides identity labels, the dataset includes body part masks and timestamps, and it comes along with two ecologically motivated splits.
We used the \textit{closed-set} split, where training data originates from encounters prior to the year 2020; newer observations form the test set.
Unlike with lynx data, images in the Loggerhead sea turtle dataset are captured underwater by a moving camera (see \Cref{fig:turtles-qualitative}). The precise location of image acquisition is not known. However, the image capture date is important because it reflects the seasonality. We present this dataset as an example of a situation where the impact of removing the background is not obvious, yet experiments confirm its benefit.

A similar set of identities is likely to be observed within one season, but the distribution of identities might change in the next diving season the following year. We, therefore, use the Time Decay prior (see \Cref{sec:backgroundmodel}) as our background model in this dataset and set the prior hyperparameter $\beta = 3.0$
The value of $\beta$ was found empirically on the validation set; range $[1, 10]$ with step 1.
in all experiments. \vspace{-0.1cm}\\

\paragraph{Results.} Compared to the baseline method as well as to previous results~\cite{seaturtleid2022}, incorporating the Time Decay instance prior to the network calibrated using the PITS loss brings a significant improvement. More specifically, the accuracy improved by 8.6 and 9.2 percentage points compared to Adam et al. \cite{seaturtleid2022} and our baseline, respectively. Furthermore, the use of instance prior $\Psi_k$ contributed with 4\%. For more details about performance, see ~\Cref{tab:turtles}.


\begin{table}
\vspace{-5pt}
\setlength{\tabcolsep}{0.75em}
  \footnotesize
  \centering
  \vspace{-2pt}
  \caption{Caretta caretta re-identification accuracy. (Left) Adam et al.\,\cite{seaturtleid2022}. (Right) our.}
  \begin{tabular}{@{}lc|c@{}}
    \toprule
    Input & Loss  & Accuracy \\
    \midrule
    Whole image\,\cite{seaturtleid2022} & ArcFace &    17.1\% \\
    \textit{Background}\,\cite{seaturtleid2022} & ArcFace &  ~\,3.9\%  \\ 
    Foreground\,\cite{seaturtleid2022} & ArcFace & 14.3\%  \\ 
    \bottomrule
    \vspace{0.55cm}
  \end{tabular}
  \hspace{0.2cm}
    \begin{tabular}{@{}lc|c@{}}
    \toprule
    Input & Loss  & Accuracy \\
    \midrule
    Whole image  & CE &    16.5\% \\
    \textit{Background} & CE &  ~\,7.6\%  \\ 
    Foreground & CE & 21.6\%  \\ 
    Foreground & PITS &  \underline{21.7}\%  \\ 
    \midrule
    Foreground + $\Psi_k$ & PITS &  \textbf{25.7\%}  \\ 
    \bottomrule
  \end{tabular}
  \label{tab:turtles}
    \vspace{-30pt}
\end{table}

  
\subsection{Ablations}
\paragraph{Calibration.} We compare the proposed PITS loss to the standard CE loss, optionally with temperature scaling~\cite{guo2017calibration} as the post-processing step. As shown in \Cref{tab:calibrationablation}, our PITS loss consistently improves accuracy on a whole image and on an image with foreground pixels only. This effect is further enhanced when the output calibration through the PITS loss is combined with instance prior\,\,$\Psi_k$. Standard temperature scaling also works with the instance prior $\Psi_k$ and improves overall accuracy, but less than PITS.

\begin{table}[h]
  \vspace{-5pt}
  \setlength{\tabcolsep}{0.45em}
  \footnotesize
  \centering
  \caption{Calibration ablation. Accuracy (Acc) and Expected Calibration Error (ECE) on the Eurasian lynx dataset.}
  \begin{tabular}{lc|cc|cc|cc}
    \toprule
          &       & \multicolumn{6}{c}{Calibration Method} \\
     & Background& \multicolumn{2}{c}{\textit{None}} & \multicolumn{2}{c}{Temp. scaling} & \multicolumn{2}{c}{PITS} \\
     Input & model  $\Psi_k$ & Acc.$\uparrow$ & ECE$\downarrow$ & Acc.$\uparrow$ & ECE$\downarrow$ & Acc.$\uparrow$ & ECE$\downarrow$ \\
    \midrule
    Whole image & -- & 49.7\% & 35.6\% & 49.7\% & 27.4\% & 50.0\% & 36.6\%\\
    Whole image + $\Psi_k$& $\text{HL}_\theta$ & 47.8\% & 34.5\% & 52.9\% & 31.4\% & 55.5\%& 33.2\%\\
    Whole image + $\Psi_k$& $\text{ML}_\theta$ & 42.3\% & 33.2\% & 54.1\% & 30.7\%& 55.0\%& 34.5\% \\ 
    \midrule
    Foreground & -- & 49.3\% & 34.5\% & 49.3\% & 27.9\%& 52.5\%& 31.2\% \\
    Foreground + $\Psi_k$ & $\text{HL}_\theta$ & 56.7\% & 32.4\%& 54.2\% & 29.4\% & \underline{59.7}\%& \underline{28.8}\% \\
    Foreground + $\Psi_k$ & $\text{ML}_\theta$ & 57.1\% & 32.6\% & 57.2\% & 29.2\% & \textbf{60.9}\%& \textbf{26.6}\% \\ 
    \bottomrule
  \end{tabular}
  \label{tab:calibrationablation}
  \vspace{-5pt}
\end{table}

\paragraph{Camera Location Information.} One might argue that the problem at hand might be solved by simply giving the network explicit information about camera location as an additional input.
To test the effectiveness of providing explicit camera location information as an additional input, we conduct an experiment where camera locations are inpainted into images. 
Each location is represented uniquely to facilitate the network's use of this injected feature\footnote{Each camera location has its unique (generated) name and font color to ensure the network can easily exploit this artificially injected feature.} (see Fig. \ref{fig:priorinjection}).

Results in \Cref{tab:prioinjection} show a slight improvement in accuracy but a notable decrease in accuracy at new locations. Still, our method maintains a more than 8 percentage point advantage in overall accuracy and accuracy at new locations over the unpainted text. 
Injecting camera location directly into images reveals the network's focus on this information for identification. The network often prioritizes camera location pixels in challenging cases.
\begin{table}[h]
\vspace{-0.25cm}
  \footnotesize
  \setlength{\tabcolsep}{0.85em}
  \centering
   \caption{Camera location injection method. Impainting the location into the image positively impacts performance at "known" locations but decreases the performance at "unknown" locations. Our method, based on PITS and prior $\Psi_k$ -- $\text{ML}_\theta$ outperforms it by 8.8\% while increasing the accuracy at "new locations" by 5.7\% over the baseline.}
  \begin{tabular}{lcc|cc}
    \toprule
     &  &  & \multicolumn{2}{c}{Accuracy} \\
     Image & Location encoding & Loss & \textit{overall}  & \textit{new location} \\
    \midrule
    {Whole image}  & background pixels &  {CE} & {49.7\%} & {13.9\%}\\
    Foreground & none &  CE & 49.3\% & \underline{16.3}\% \\ 
    {Foreground} & inpainted text & {CE} & {\underline{52.1}\%} & {11.0\%} \\
    Foreground&  $\Psi_k$ -- $\text{ML}_\theta$ & PITS  & \textbf{60.9\%} & \textbf{19.6\%} \\ 
    \bottomrule
  \end{tabular} 
  \label{tab:prioinjection}
\end{table}
\begin{figure}[h]
\vspace{-0.9cm}
    \centering
    \setlength\tabcolsep{1pt}
    \begin{tabular}{ccc}
         \includegraphics[height=2.8cm, width=3.8cm]{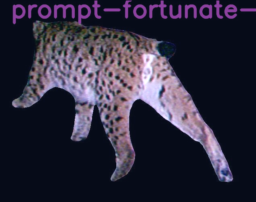} & 
         \includegraphics[height=2.8cm, width=3.8cm]{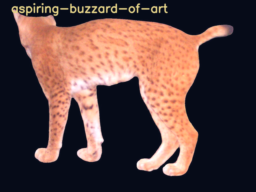} & 
         \includegraphics[height=2.8cm, width=3.8cm]{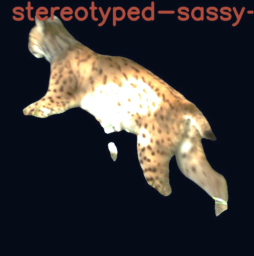} \vspace{-1px}\\ 
         \includegraphics[height=2.8cm, width=3.8cm]{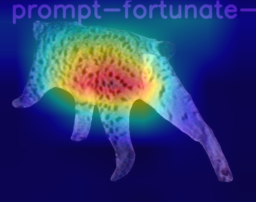} & \includegraphics[height=2.8cm, width=3.8cm]{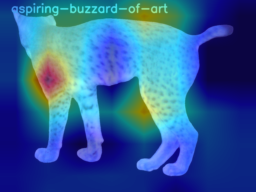} & \hspace{-3px}\includegraphics[height=2.8cm, width=3.8cm]{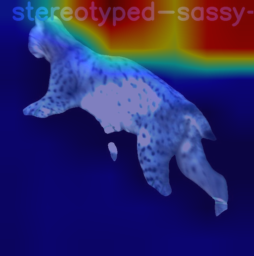} \\ 
    \end{tabular}
    \caption{Impainting location ablation. In some cases, \textit{location} plays no role in identification (\textit{left}), often the \textit{location} is combined with individuals' appearance (\textit{middle}), and in challenging cases, only the \textit{location} is used to identify the individual (\textit{right}).}    
    \label{fig:priorinjection}
    \vspace{-0.15cm}
\end{figure}
\newpage
\paragraph{Backbone network.} Last but not least, we present results with different backbones. \Cref{tab:architectures} shows that the proposed PITS loss consistently improves accuracy for all tested architectures when compared to the standard CE loss and that combining the PITS loss with Identity Prior $\Psi_k$ achieves the best accuracy for all backbone architectures. The absolute accuracy of individual architectures seems to be mostly affected by the relatively small number of samples in the training dataset, and therefore larger models seem to suffer from overfitting.

\begin{table}[h]
  \setlength{\tabcolsep}{0.85em}
  \centering
  \caption{Backbone network ablation on the Eurasian lynx dataset. With all the architectures, the PITS loss with Identity Prior $\Psi_k$ maintains the constant gain of around 10 percentage points over the baseline.}
  \begin{tabular}{lc|ccc}
    \toprule
    Architecture & Parameters & CE & PITS & PITS + $\Psi_k$ \\
    \midrule
    Swin-T~\cite{liu2021swin} & 27.6M & 37.9\% & \underline{43.0}\% & \textbf{49.9}\% \\    
    ResNet-50~\cite{he2016deep} & 23.6M & 41.2\% & \underline{45.1}\% & \textbf{51.3}\%\\    
    ResNeXt-50~\cite{xie2017aggregated} & 23.1M & 44.9\% & \underline{48.4}\% & \textbf{56.4}\%\\    
    EfficientNetV2 S~\cite{tan2021efficientnetv2} & 20.3M & 48.5\% & \underline{53.4}\% & \textbf{59.9}\% \\    
    EfficientNetV2 M~\cite{tan2021efficientnetv2} & 52.9M & 49.1\% & \underline{53.1}\% & \textbf{60.7}\% \\    
    EfficientNet B3~\cite{tan2019efficientnet} & 10.8M & 49.7\% & \underline{52.5}\% & \textbf{60.9}\% \\      
    \bottomrule
  \end{tabular}
  \label{tab:architectures}
  \vspace{-0.5cm}
\end{table}

\section{Conclusion}
\label{sec:conclusion}

A new method for robustly exploiting the background and the foreground in visual identification was proposed.
By separating the foreground and the background information processing in training, and then combining the predictions in a probabilistic justified way, the relative error of the baseline method on two public wildlife re-identification datasets of endangered species (e.g., \textit{Lynx lynx} and \textit{Caretta caretta}) was reduced by 22.3\% and 10.4\% respectively (i.e. {accuracy was increased by more than 11 and 8 percentage points respectively). When considering cases where an object 
appears in new locations, the recognition accuracy almost doubled over standard methods. 

The main limitation is the reliance on accurate foreground-background segmentation and the dependency on task-specific priors, which have to be crafted manually, and which might differ from species to species. Another limitation is the assumption less frequent identities are harder for the network to learn because of the lack of training samples, which might not necessarily always be the case, as some distinct individuals might be easily distinguishable even from a small set of training samples. \\

\paragraph{Acknowledgement.}
The authors were supported by the Technology Agency of the Czech Republic, project No. SS05010008.
Computational resources were provided by the e-INFRA CZ project (ID:90254), supported by the Ministry of Education, Youth and Sports of the Czech Republic.
We also thank Friends of the Earth, the Czech Republic, for providing the \textit{Lynx lynx} data.

\bibliographystyle{splncs04}
\bibliography{main}

\end{document}